\documentclass{nle}

\usepackage{amsmath}
\usepackage{graphicx}
\usepackage{natbib}
\ifpdf%
\usepackage{epstopdf}%
\else%
\fi
\usepackage{multirow}

\usepackage{enumitem}
\usepackage{multicol}
\usepackage{float}
\usepackage{breqn}
\usepackage{arydshln}
\usepackage[hyphens]{url}
\usepackage{color, colortbl}
\usepackage{array, makecell}
\usepackage{sidecap}
\usepackage{subfig}
\usepackage{amsfonts}

\definecolor{LightRed}{rgb}{0.99,0.8,0.8}
\definecolor{LightGreen}{rgb}{0.6,0.9,0.6}

\newcommand{\model}{\texttt{FiADD}}
\newcommand{\modelfull}{\textbf{F}ocused \textbf{I}nferential \textbf{A}daptive \textbf{D}ensity \textbf{D}iscrimination}
\newcommand{\abuse}{\texttt{AbuseEval}}
\newcommand{\latent}{\texttt{LatentHatred}}
\newcommand{\gab}{\texttt{ImpGab}}
\newcommand{\ADD}{ADD}
\newcommand{\ADDFOC}{$ADD^{foc}$}
\newcommand{\ADDINF}{$ADD^{inf}$}
\newcommand{\ADDINFFOC}{$ADD^{inf + foc}$}

\newcommand{\ADDFull}{\texttt{Adaptive Density Discrimination}}

\begin{document}
\label{firstpage}

\lefttitle{FiADD}
\righttitle{Natural Language Engineering}

\papertitle{Article}

\jnlPage{1}{00}
\jnlDoiYr{2019}
\doival{10.1017/xxxxx}

\title{Focal Inferential Infusion Coupled with Tractable Density Discrimination for Implicit Hate Detection}

\begin{authgrp}
\author{Sarah Masud$^{*1}$}
\author{Ashutosh Bajpai$^{*2,3}$}
\author{ Tanmoy Chakraborty$^{2,\#}$}
\affiliation{$^1$ Indraprastha Institute of Information Technology, Delhi,  New Delhi, India\\
$^2$ Indian Institute of Technology Delhi, New Delhi, India \\
$^3$ Wipro Research, India \\
\email{sarahm@iiitd.ac.in, ashutoshbajpai.ma.gre@gmail.com, tanchak@iitd.ac.in}
}
\affiliation{$^*$ Equal Contribution}
\affiliation{$^\#$ Corresponding author}
\end{authgrp}


\begin{abstract}
Although pretrained large language models (PLMs) have achieved state-of-the-art on many natural language processing (NLP) tasks, they lack an understanding of subtle expressions of {\em implicit} hate speech. Various attempts have been made to enhance the detection of implicit hate by augmenting external context or enforcing label separation via distance-based metrics. Combining these two approaches, we introduce \model, a novel \modelfull\ framework. \model\ enhances the PLM finetuning pipeline by bringing the surface form/meaning of an implicit hate speech closer to its implied form while increasing the inter-cluster distance among various labels. We test \model\ on three implicit hate datasets and observe significant improvement in the two-way and three-way hate classification tasks. We further experiment on the generalizability of \model\ on three other tasks, detecting sarcasm, irony, and stance, in which surface and implied forms differ, and observe similar performance improvements. Consequently, we analyze the generated latent space to understand its evolution under \model, which corroborates the advantage of employing \model\ for implicit hate speech detection.
\end{abstract}

\maketitle

\section{Introduction}
The Internet has led to a proliferation of hateful content \citep{Suler2004TheOD}. However, what can be considered hate speech is subjective \citep{10.1145/3479158,10.1145/3363565}. According to the United Nations\footnote{\url{https://www.un.org/en/hate-speech/understanding-hate-speech/what-is-hate-speech}}, \textit{hate speech is any form of discriminatory content that targets or stereotypes a group or an individual based on identity traits.}  
\textcolor{black}{In order to assist content moderators, practitioners are now looking into automated hate speech detection techniques. The paradigm that is currently being adopted is finetuning a pretrained language model (PLM) for hate speech detection. Akin to any supervised classification task, the first step is the curation of hateful instances.} While instances of online hate speech have increased, they still form a small part of the overall content on the Web. For example, on platforms like Twitter, the ratio of hate/non-hate posts curated from the data stream is 1:10 \citep{10.1145/3580305.3599896}. Thus, data curators often employ lexicons and identity slurs to increase the coverage of hateful content\footnote{\color{red}{\textbf{Disclaimer:} The paper contains samples of hate speech, which are only included for contextual understanding.}}. While this increases the number of explicit samples, it comes at the cost of capturing fewer instances of implied/non-explicit hatred \citep{davidson2017automated, Silva_Mondal_Correa_Benevenuto_Weber_2021}. \textcolor{black}{This skewness in the number of implicit samples contributes to less information being available for the models to learn from.} Among the myriad datasets on hate speech \citep{Vidgen2020,Poletto2021} in English, only a few \citep{latent, abuse, gab} have annotations for ``implicit" hate. 

\textbf{Why is implicit hate hard to detect?} It has been observed that classifiers can work effectively with direct markers of hate \citep{lin-2022-leveraging,10.1145/3582568}, \emph{a.k.a} explicit hate. The behavior stems from the data distribution since slurs are more likely to occur in hateful samples than in neutral ones. On the other hand, implicit hate on the surface appears lexically and semantically closer to statements that are non-hate/neutral. Inferring the underlying stereotype and implied hatefulness in an implicit post requires a combination of multi-hop reasoning with sufficient cultural reference and world knowledge. Existing research has established that even the most sophisticated systems like ChatGPT perform poorly in case of implicit hate detection \citep{yadav2024toxbartleveragingtoxicityattributes}. 

At the distribution level, the aim is to bring the surface meaning closer to its implied meaning, i.e., what is said vs. what is intended \citep{latent, lin-2022-leveraging}. One way to reduce the misclassification of implicit hate is to manipulate the inter-cluster latent space via contrastive or exemplar sampling \citep{kim-etal-2022-generalizable}. Contrastive loss similar to cross-entropy operates in a per-sample setting \citep{1467314}, leading to sub-optimal separation among classes \citep{pmlr-v48-liud16}. Another technique is to infuse external knowledge. However, without explicit hate markers, providing external knowledge increases \textcolor{black}{the noise in the input signal \citep{lin-2022-leveraging, yadav2024toxbartleveragingtoxicityattributes}.} 

 \begin{figure*}[!t]
\centering
\includegraphics[width=0.65\columnwidth]{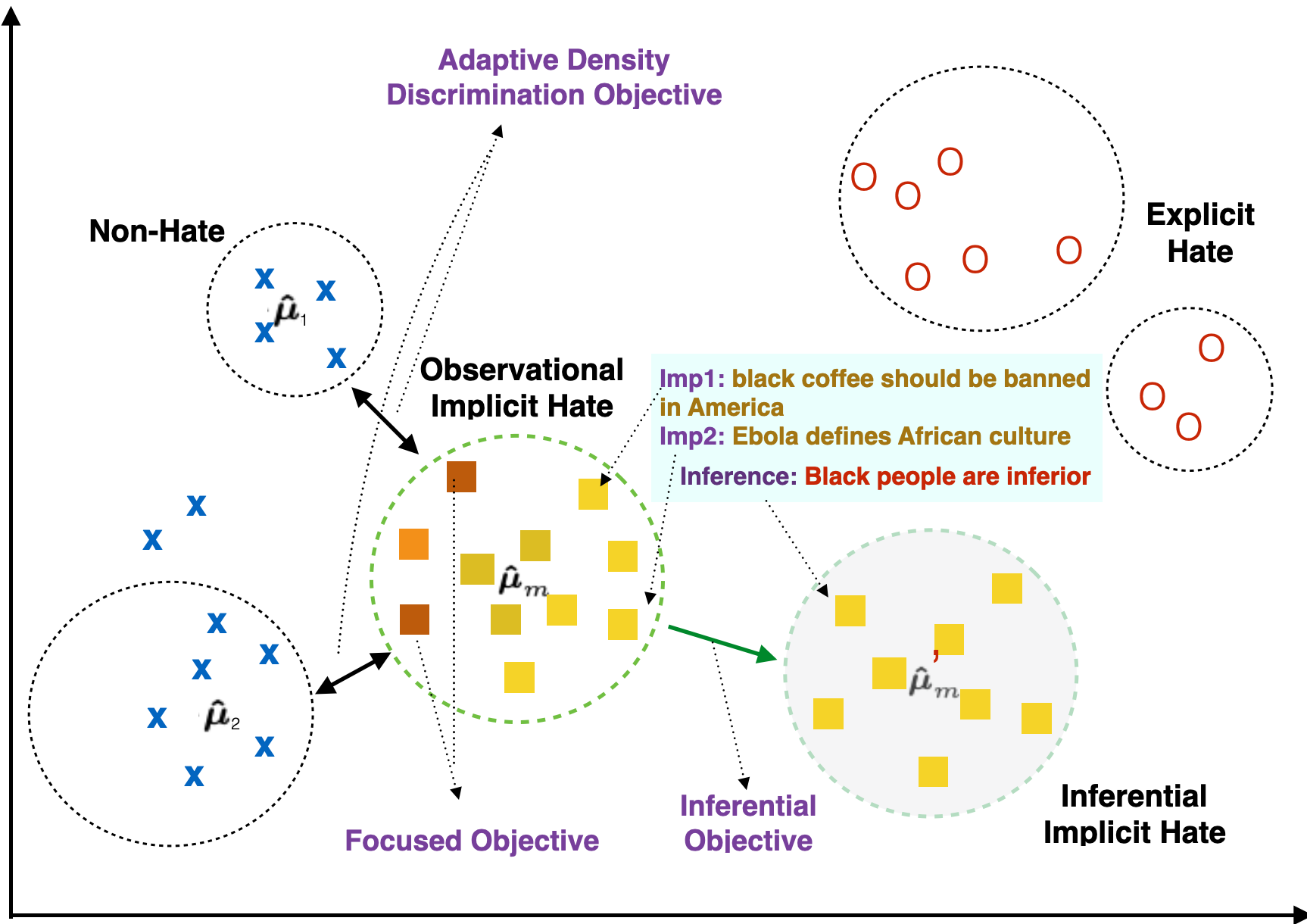}
\caption{\textcolor{black}{The three objectives of \model\ as applied to implicit hate detection are (a) adaptive density discrimination, (b) higher penalty on boundary samples, and (c) bringing the surface and semantic form of the implicit hate closer.}}
\label{fig:intution}
\vspace{-4mm}
\end{figure*}

\textbf{Proposed framework.}
\textcolor{black}{In this work, we examine a framework for overcoming these two drawbacks. As an alternative to the per-sample contrastive approach} in computer vision tasks, Adaptive Density Discrimination (\ADD) \textit{a.k.a} Magnet Loss \citep{magnetloss} has been proposed. \ADD\ does not employ the most positively and negatively matching sample; instead, it exploits the local neighborhood to balance the inter-class similarity and variability. Extensive literature \citep{magnetloss, Deng_2019_CVPR,NIPS2017_cb8da676} has established the efficacy and superiority of \ADD\ for computer vision tasks over contrastive settings. We hypothesize that its advantage can be extended to NLP and attempt to establish the same in this work. 

For our use case, \ADD\ can help improve the regional boundaries around implicit and non-hate samples that lie close. \textcolor{black}{However, simply employing \ADD\ in a three-way classification of implicit, explicit, and non-hate will not yield the desired results due to the semantic and lexical similarity of implicit with non-hate. We, thus, introduce external context for implicit hate samples to bring them closer to their intended meaning \citep{kim-etal-2022-generalizable}, facilitating them to be sufficiently discriminated.} To this end, we employ implied/descriptive phrases instead of knowledge tuples or Wikipedia summaries based on empirical findings \citep{lin-2022-leveraging,yadav2024toxbartleveragingtoxicityattributes} that the latter tend to be noisy if the tuples are not directly aligned with the entities in the input statement. \textcolor{black}{As outlined in Figure \ref{fig:intution}, our proposed pipeline \modelfull\ (\model) improves the detection of implicit hate by employing distance-metric learning to set apart the class distributions in conjunction with reducing the latent space between the implied context and implicit hate. The dual nature of the loss function is aided by non-uniform weightage, with a focus on penalizing samples near the discriminant boundary.} 

\textcolor{black}{Throught extensive experiments, we observe that \model\ variants improve overall as well as implicit class macro-F1 for \latent\ \citep{latent}, \gab\ \citep{gab}, and \abuse\ \citep{abuse} datasets. Our experimental results further suggest that our framework can generalize to other tasks where surface and implied meanings differ, such as humor \citep{Labadie_TamayoChulviRosso2023}, sarcasm \citep{abu-farha-etal-2022-SemEval,10.1007/978-3-031-42448-9_4}, irony \citep{van-hee-etal-2018-SemEval}, stance \citep{mohammad-etal-2016-SemEval}, etc.} To establish that our results are not BERT \citep{devlin-etal-2019-bert} specific, we also experiment with HateBERT \citep{caselli-etal-2021-HateBERT}, XLM \citep{chi-etal-2022-xlm}, and LSTM \citep{Hochreiter1997}.

\textbf{Contributions.} In short, we make the following contributions through this study\footnote{\textbf{Reproducibility.} Sample code and dataset are available at \url{https://github.com/LCS2-IIITD/FIADD}.}:
\begin{itemize}[noitemsep,nolistsep,topsep=0pt,leftmargin=1em]
    \item \textcolor{black}{We perform a thorough literature survey of the implicit hate speech datasets. For the datasets employed in this study,} we establish the closeness of implicitly hateful samples from non-hateful ones (Section \ref{sec:motivation}) and \textcolor{black}{use it to motivate our model design (Figure \ref{fig:intution}).}
\item \textcolor{black}{We adopt \ADD\ for the NLP setting and employ it to propose \modelfull\ (\model). The variants of the proposed setup allow it to be used as a pluggable unit for the PLM finetuning pipeline for the task of hate speech detection as well as other implicit text-based tasks (Section \ref{sec:framework}).} 
\item We manually generate implied explanations/descriptions for $798$ and  $404$ implicit hate samples for \abuse\ and \gab, respectively. \textcolor{black}These annotations contribute to corpora of unmasking implicit hate (Section \ref{sec:exp}).
\item Our exhaustive experiments, analyses, and ablations highlight how \model\ compares with the cross-entropy loss on three hate speech datasets. We also extend our analysis to three other SemEval tasks to demonstrate the model's generalizability (Section \ref{sec:results}). 
    \item We perform an analysis to assess how the latent space evolves under \model\ (Sections \ref{sec:analysis}).
\end{itemize}

\textbf{Research scope and social impact.} Early detection of implicit hate will help reduce the psychological burden on the target groups, prevent conversation threads from turning more intense, and also assist in counter-hate speech generation. It is imperative to note the limitations of PLMs in understanding implicit hate speech. We attempt to overcome this by incorporating latent space alignment of surface and implied context. However, PLMs cannot replace human content moderators and can only be assistive.  

\section{Related Work}
\label{sec:lit}
Given that this study proposes a distance-based objective function primarily for implicit hate detection, \textcolor{black}{the literature survey focuses on three main aspects --- (i) implicit hate datasets, (ii) implicit hate detection, and (ii) improvement in the classification tasks via distance-based metrics.} \textcolor{black}{To determine the relevant literature for implicit hate within the vast hate speech literature,} we make use of the up-to-date hate speech corpus\footnote{https://hatespeechdata.com/} \citep{Vidgen2020} as well ACL Anthology. The keywords used to search for relevant literature on the two corpora were `implicit' and `implicit hate,' respectively. 

\textcolor{black}{\textbf{Implicit hate datasets.} The task of classifying hateful texts has led to an avalanche of hate detection datasets and models \citep{schmidt-wiegand-2017-survey, Vidgen2020,10.1145/3290838}.} Before discussing the literature, it is imperative to point out that issues with generalizable \citep{Yin2021}, biasing \citep{10.1145/3580494, 10.1145/3479158}, adversary \citep{masud2024hatepersonifiedinvestigatingrole}, outdated benchmarks \citep{masud-etal-2024-probing} are prevalent for hate speech detection at large, and forms an active area of research. 

\textcolor{black}{Focusing on implicit hate datasets, we searched the hate speech database \citep{Vidgen2020} with the keyword `implicit' as an indicator of whether the label set contains `implicit' labels and obtained $4$ results. DALC \citep{caselli-etal-2021-dalc} is a Dutch dataset consisting of $8k$ tweets curated from Twitter, labeled for the level of explicitness as well as the target of hate. Meanwhile, ConvAbuse consists of 4k English samples obtained from in-the-wild human-AI conversations with AI chatbots. Each conversation is marked for the degree of abuse (1 to -3) and directness (explicit or implicit). The other two datasets are also in English. \abuse\ \citep{abuse} is $14k$, Twitter labeled for `abusiveness' and `explicitness.' On the other hand, \gab\ \citep{gab} consists of 27k posts from Gab, which contain a hierarchy of annotations about the type and target of hate.}   

\textcolor{black}{Meanwhile, from the ACL Anthology (we looked at the results from the first two pages out of 10), we discovered four more datasets. \latent\ is the most extensive and most widely used implicit hate speech dataset. It consists of $21k$ Twitter samples labeled for implicit hate as well as $6$ additional sub-categories of implicitness. It also contains free-text human annotations explaining the implied meaning behind the implicit posts. Along similar lines, SBIC \citep{ocampo-etal-2023-depth} is also a collection of $44k$ implicit posts curated from online platforms with human-annotated explainations. However, unlike complete sentences in \latent\, SBIC focuses on single-phrased explainations. Further, SBIC does not have a direct marker for the explicitness of the post, and by default, all posts are implicit. For specific target groups and types of hate speech, such as sexism \citep{kirk-etal-2023-semeval} or xenophobia against immigrants \citep{app11083610}, researchers have also explored imploying multiple-level annotations as a means of obtaining granular label spans as explainations for the hateful instance. It serves as an alternative to free-text annotations, allowing for more structured and linguistic \textcolor{black}{analysis \citep{PLN6481}} of implicitness. Further, building upon the multimodal hate meme dataset MMHS150K \citep{Gomez_2020_WACV}, proposed a multimodal implied hate dataset \citep{botelho-etal-2021-deciphering} with the different types of implicitness occurring as a combination of the text and image.}

More recently, the ISHate \citep{ocampo-etal-2023-depth} dataset has been curated by combining existing hate speech and counter-hate speech datasets and relabelling the samples for explicit-implicit markers, consisting of $30k$ samples labeled as explicit, implicit, subtle, or non-hate. It is interesting to note that in their analysis, the authors do not showcase how the different datasets interact with each other in the latent space. We hypothesize that the performance improvements in hate detection are obtained not as a result of modeling but due to the fact that these samples are obtained from distinct datasets, i.e., distinct distributions. For example, counter-hate datasets do not contribute to the non-hate class. Meanwhile, the majority of implicit hate samples come from \latent\ and ToxiGen \citep{hartvigsen-etal-2022-toxigen}. The latter is a curation of around 1M toxic and implicit statements obtained by controlled generation.

\textbf{Modeling implicit hate speech in NLP.} \textcolor{black}{Despite a large body of hate speech benchmarks, the majority of datasets fail to demark implicit hate. Even during the annotation process, fine-grained variants of offensiveness \cite{founta2018large, 10.1145/3580305.3599896, kirk-etal-2023-semeval} like abuse, provocation, sexism, etc, are favored over the nature of hate, i.e., explicit vs implicit. As the annotation schemas have a direct impact on the downstream tasks \citep{rottger-etal-2022-two}, the common vouge of binary hate speech classification, while easier to annotate and model, focuses on explicit forms of hate. It also comes at the cost of not analyzing the erroneous cases where implicit hate is classified as neutral content. This further motivates us to examine the role of PLMs in three-way classification in this work.} 

\textcolor{black}{Given the skewness in the number of implicit hate samples in a three-way classification setup, data augmentation techniques have been explored.} For example, \citep{ocampo-etal-2023-depth} employed multiple data augmentation \textcolor{black}{like substitution, back translation, etc., and observed that only when multiple techniques are combined did they surpass the finetuned HateBERT in performance.} Adversarial data collection \citep{ocampo-etal-2023-playing} and LLM-prompting \citep{kim-etal-2023-conprompt} have also been explored for augmenting and improving implicit hate detection. 

\textcolor{black}{Language models are being employed not only to augment the implicit hate corpora but also to detect hate \citep{ghosh-senapati-2022-hate, plaza-del-arco-etal-2023-respectful}.} With the recent trend of prompting generative large language models (LLMs), hate speech detection is now being evaluated under zero-shot \citep{plaza-del-arco-etal-2023-respectful, nozza-2021-exposing, masud2024hatepersonifiedinvestigatingrole} and few-shot \citep{chiu2022detecting} settings as well. \textcolor{black}{An examination of the hate detection techniques under fine-grained hate speech detection has revealed that traditional models, either statistical \citep{davidson2017automated, waseem-hovy-2016-hateful} or deep learning-based \citep{founta2018unified, Badjatiya}, are characterized by a low recall for hateful samples \citep{10.1145/3580305.3599896}. To increase the information gained from the implicit samples, researchers are now leveraging external context.} 

Studies have mainly explored the infusion of external context in the form of knowledge entities, either in the form of knowledge-graph (KG) tuples \citep{latent} or Wikipedia summaries \citep{lin-2022-leveraging}. However, both works have observed that knowledge infusion at the input level lowered the performance on fine-grained implicit categories. An examination of the quality of knowledge \textcolor{black}{tuple \citep{yadav2024toxbartleveragingtoxicityattributes}} infusion for implicit hate reveals that KG tuples fail to enlist information that directly connects with the implicit entities, acting more as noise than information. Apart from textual features, social media platform-specific features like user metadata, user network, and conversation thread/timeline can also be employed to improve the detection of hate and capture implicitness in long-range contexts \citep{ghosh-etal-2023-cosyn}. However, such features are platform-specific, complex to curate, and resource-intensive to operate (in terms of storage and memory to train network embeddings). From the latent space perspective, researchers have explored how the infusion of a common target group can bring explicit and implicit samples closer \citep{ocampo-etal-2023-unmasking}, aiding in the detection of the latter. While the idea is intuitive since implicit hate and explicit slurs are specific to a target group, here, the extent of overlap in the case of multiple target groups or intersectional identities is not adequately addressed.

\textbf{Distance-metric learning.} \textcolor{black}{Akin to supervised classification task in NLP, all the setups reviewed so far use an encoder-only BERT-based + Cross entropy (CE) for finetuning. Therefore, in our study, BERT+CE acts as a baseline. Despite its popularity, CE's impact on the inter/intra-class clusters is sub-optimal \citep{pmlr-v48-liud16}.} Since classification tasks can be modeled as obtaining distant clusters per class, one can exploit clustering and distance-metric approaches to enhance the boundary among the labels, leading to improved classification performance. Distance-metric learning-based methods employ either deep divergence via distribution \citep{pmlr-v119-cilingir20a} or point-wise norm \citep{1467314}. The most popular deep metric learning is the contrastive loss family \citep{1467314, Schroff_2015_CVPR,8099628}. In order to improve upon the cross entropy loss (CE) and benefit from the one-to-one mapping of the implicit hate and its implied meaning, contrastive learning has been explored \citep{kim-etal-2022-generalizable}, which has only provided slight improvement. 

However, like cross-entropy, contrastive loss operates on a per-sample basis; \textcolor{black}{even when considering positive and negative exemplars, they are curated on a per-sample basis.} Clustering-inspired methods \citep{magnetloss, 8099720} have sought to overcome this issue by focusing on subclusters per class. \ADD\ {\em a.k.a} magnet loss \citep{magnetloss} specifically lends a good starting point to operate the shift in the inter-cluster distance to extend to our use case. Based on the fact that \ADD\ has surpassed contrastive losses in other tasks, we use \ADD\ as a starting point and improve upon its formulation for implicit detection. \textcolor{black}{As the current \ADD\ setup fails to account for the implied meaning, we infuse external information into the latent space as an implied/inferential cluster.}

\begin{SCtable}
\caption{The L1 inter-cluster distances between neutral (N) and explicit hate (E)), as well as non-hate and implicit hate (I) samples based on \textcolor{black}{ALD and ACLD.}}
\label{tab:intercluster}
\resizebox{0.45\columnwidth}{!}{
\begin{tabular}{|c|c|c|c|c|}
\hline
\multirow{2}{*}{Dataset} & \multicolumn{2}{c|}{ALD} & \multicolumn{2}{c|}{ACLD} \\ \cline{2-5}
                         & N-E        & N-I        & N-E         & N-I        \\ \hline
\abuse\    & 132.73     & 128.81     & 96.69       & 94.94      \\\hline
\gab\ & 132.32     & 131.62     & 97.62       & 97.08      \\\hline
\latent\   &    125.12        &   122.59         &       90.86      &  90.69  \\\hline       
\end{tabular}
}
\vspace{-5mm}
\end{SCtable}

\section{Intuition and Background}
\label{sec:motivation}
This section attempts to establish the need for distance-metric-based learning for the task of hate speech detection. \textcolor{black}{Inspired by our initial experiments, we provide an intuition for \ADD.}

\textbf{Hypothesis.} A manual inspection of the hate speech datasets reveals that non-hate is closer to implicit hate than explicit hate. We, thus, measure the inter-cluster distance between non-hate and implicit hate compared to non-hate and explicit hate. 

\textbf{Setup.} For three implicit hate speech datasets, i.e., \latent\, \gab\, and \abuse, we embed all the samples for a dataset in the latent space using $768$ dimensional CLS embedding from BERT. The embeddings are not finetuned on any dataset or task related to hate speech so as to reduce the impact of confounding variables. We then consider 3 clusters directly adopted from the implicit, explicit, and non-hate classes and record the pairwise average linkage distance (ALD) and average centroid linkage distance (ACLD) among these clusters. 

As the name suggests, for \textcolor{black}{ACLD}, \textcolor{black}{we first obtain the embedding for the center of each cluster as a central tendency (mean or median) of all its representative samples and then compute the distance between the centers. This distance indicates the overall closeness of the two centers, which, in our case, measures the extent of similarity between the two classes.} We also assess the latent space more granularly via \textcolor{black}{ALD.} In ALD, the distance between two clusters is obtained as the average distance between all possible pairs of samples where each element of the pair comes from a distinct group. \textcolor{black}{It allows for a more fine-grained evaluation of the latent space, as not all data points are equidistant from each other or their respective centers.} \textcolor{black}{Formally,} consider a system with E ($\mathbb{R}\in d$) points, where each point ($e_i$) belongs to one of the N clusters $c^n(e_i)$ and $\mu^n = \frac{1}{|c^n|}\sum_{e_i \in c^n}e_i$ is the cluster center. For clusters $a$ and $b$, $ACLD^{a,b}=dist(\mu^a,\mu^b)$. Meanwhile,  $ALD^{a,b}=\frac{1}{|c^a| * |c^b|}\sum_{e_i \in c^a \& e_j \in c^b}dist(e^i,e^j)$ where ($c^a(e_i)\neq c^b(e_j)$). 

\textcolor{black}{The intuition behind using both ACLD and ALD stems from the fact that online hate speech is part of the larger discourse on the Web.} Thus, it is possible that at the level of individual datapoint labeling an isolated instance as hateful is hard. Furthermore, some implicit samples may be closer to the explicit hate samples in terms of lexicon or semantics. On the other hand, it is also possible for some non-hate samples to contain slurs that are commonplace and context-specific but not objectionable within the community \citep{diaz-etal-2022-accounting,rottger-etal-2021-hatecheck}. \textcolor{black}{ACLD and ALD allow us to capture these dynamics at a macroscopic and a microscopic level.} 

\textbf{Observation.} \textcolor{black}{From Table \ref{tab:intercluster}, we observe that under both ALD and ACLD, non-hate is closer to implicit samples. As expected, ALD shows more variability than ACLD. It follows from the fact that the mere presence of a keyword/lexicon does not render a sample as hateful.}

\textcolor{black}{Stemming from these observations, we see a clear advantage of employing a distance-metric approach that can exploit the granular variability in the latent space.} \ADDFull\ (\ADD) based clustering loss, which optimizes the inter and intra-clustering around the local neighborhood, directly maps to our problem of regional variability among the hateful and non-hateful samples. Further, our observations motivate the penalization of samples closer to the boundary responsible for increasing variability. The proposed model, as motivated by our empirical observations, is outlined in Figures \ref{fig:intution} and \ref{fig:modelarchitecture}.

\subsection{\textcolor{black}{Background on Adaptive Density Discrimination.}}
\textcolor{black}{Here, we briefly outline \ADDFull\ (\ADD), which forms the backbone of our proposed framework.} \ADD\ is a clustering-based distance-metric. It evaluates the local neighborhood or clusters among the samples after each training iteration. At each epoch, after the training samples have been encoded into vector space, \ADD\ clusters all data points within a class into $K$ representative local groups via K-means clustering. \textcolor{black}{The subclusters within a class help} capture the inter/intra-label similarity \textcolor{black}{around} the local neighborhood. \textcolor{black}{If there are $N$ classes, then each training sample will belong to one of the $N*K$ subclusters.} 

Given that mapping and tracking distances among all N*K groups is computationally expensive, \ADD\ randomly selects a reference/seed cluster $I_s^c$ representing class $C$ and then picks $M$ imposter clusters from local neighborhood $I_{s1}^{c'}, \ldots, I_{sm}^{c'}$ but from disparate classes ($ c\not= c'$) based on their proximity to seed cluster. \textcolor{black}{To understand the concept of seed and imposter cluster better, consider the three-way hate speech classification task with implicit, explicit, and non-hate labels. As we aim to distinguish implicit hate speech better,} we select one of the implicit hate subclusters as the seed. Consequently, the imposter clusters will be from explicit hate or non-hate, where implicit hate can be misclassified. \ADD\ then samples $D$ points uniformly at random from each sample cluster. For the $d^{th}$ data point in $m^{th}$ cluster, $r_d^m$ is its encoded vector representation, with $C(.)$ representing the class for the sample under consideration. 
Subsequently, $\mu^m = \frac{1}{D}\sum_{d=1}^{D}r_d^m$ acts the mean representation of $m^{th}$ cluster. Here, \ADD\  applies Equation \ref{eq:magnet_single} to discriminate the local distribution around a point:
\begin{equation}
\label{eq:magnet_single}
p^{ADD}(r_d^m) = \frac{e^{-\frac{1}{2\sigma^2} \left\|r_d^m - \mu^m  \right\|_2^2 - \alpha}}{{\sum_{\mu^o:C(\mu^o)\neq C(r_m^d)}  -\frac{1}{2\sigma^2} \left\|r_d^m - \mu^o  \right\|_2^2 }} 
\end{equation}
Here, $\alpha$ is a scalar margin for the cluster separation gap. The variance of all samples away from their respective centers is approximated via $\sigma^2 =\frac{1}{MD-1}\sum_{m=1}^{M}\sum_{d=1}^{D}\left\|r_d^m - \mu^m  \right\|_2^2$.

\textcolor{black}{After each iteration, as the embedding space gets updated, so does each of the subclusters; this lends to a dynamic nature to \ADD. It allows for the selection of random subclusters and data points after each iteration. The overall loss is computed via Equation \ref{eq:mag_org_sample}.} 
\begin{equation}
\label{eq:mag_org_sample}
\ell(\Theta) = \frac{1}{MD}\sum_{m=1}^{M}\sum_{d=1}^{D}-\log{p^{ADD}}(r_d^m)
\end{equation}
\section{Proposed Method}
\label{sec:framework}
\begin{figure*}[!t]
\centering
\includegraphics[width=\textwidth]{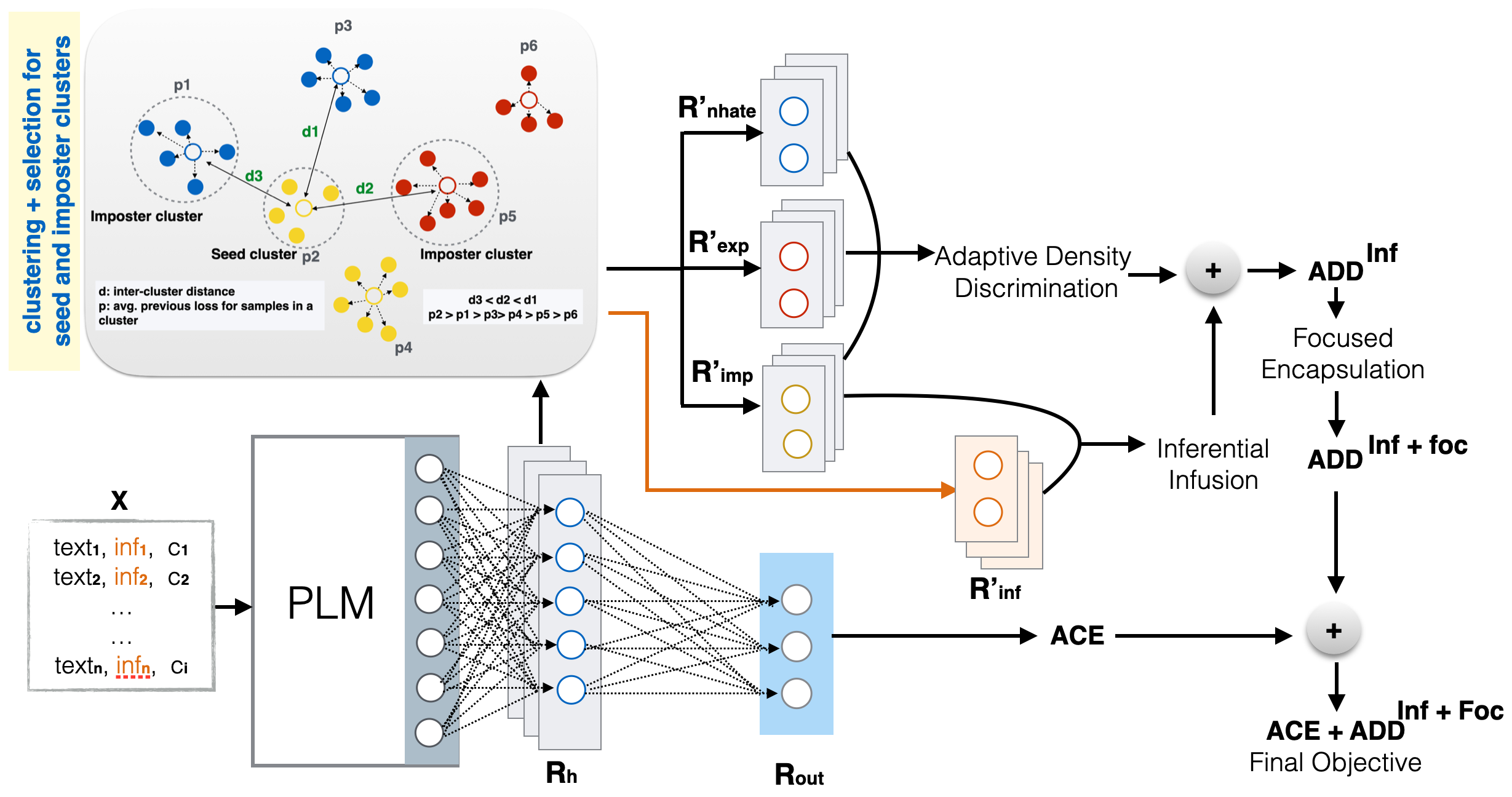}
\caption{\textcolor{black}{The architecture of \model. Input X is a set of texts, implied annotations (only for implicit class), and class labels. PLM: pretrained language model (frozen). ${R'}_{nhate}$, ${R'}_{exp}$ and ${R'}_{imp}$ are the representatives for seed and imposter clusters of non-hate, explicit, and implicit, respectively. ${R'}_{inf}$ represents inferential meaning for corresponding  ${R'}_{imp}$. ACE is alpha cross-entropy, and $ADD^{Inf+foc}$ is the adaptive density discriminator with inferential + focal objective.}}
\label{fig:modelarchitecture}
\vspace{-5mm}
\end{figure*}

\textcolor{black}{The proposed \modelfull\ (\model) framework consists of a standard finetuning pipeline with encoder-only PLM followed by a projection layer $R_h$ and a classification head (CH). To reduce the distance between the implicit hate (imp) and implied clusters (inf), \model\ measures the average distance of implicit points from implied meaning as a ratio of its distance to explicit and non-hate subspaces. During the PLM finetuning, our setup combines with cross-entropy loss to improve the detection of hate.} An overview of \model's architecture is reflected in Figure \ref{fig:modelarchitecture}. For each training instance $(x_d, y_d) \in X$, with $x_d$ input and $y_d$ label, $x_p=PLM(x_d)$ is the encoded representation obtained from the PLM. The encodings are projected to obtain $r_d = R_h(x_p)$. Here $x_p \in \mathbb{R}^{768}$ and $r_d \in \mathbb{R}^{128}$ as $r_d \ll x_d$ allows for faster clustering.

\textcolor{black}{\textbf{Novel component: inferential infusion.}} As each output label $y_d$ belongs to one of the distinct classes ($c_i \in C$), we employ the respective embeddings $r_d$ and offline K-means algorithm to obtain $K$ subclusters per class. For implicit hate samples, the latent representation of their implied/inferential counterparts $\tilde{x_d}$ is denoted as $\tilde{r_d} = R_h(\tilde{x_d})$. If $r_1^m, \ldots, r_d^m$ are representations for $D$ samples of $m^{th}$ implicit cluster, then $\tilde{r}_1^m,\ldots, \tilde{r}_d^m$ represent their respective inferential forms. The updated \textbf{I}nferential \textbf{A}daptive \textbf{D}ensity \textbf{D}iscrimination (\ADDINF) help reduce the distance between $(r_d,\tilde{r_d})$ for implicit hate samples via  Equation \ref{eq:mag_imp_sample}. 
\begin{equation}
\label{eq:mag_imp_sample}
p^{ADD^{inf}}(r_d^m) = \frac{e^{-\frac{1}{2\sigma^2} \|r_d^m - \mu^m \|_2^2 - \alpha } + 
e^{-\frac{1}{2\tilde{\sigma^2}} \|r_d^m - \tilde{\mu^m} \|_2^2 - \alpha}}{{\sum_{\mu^o:C(\mu^o)\neq C(r_m^d)}^{}  -\frac{1}{2\sigma^2} \left \|r_d^m - \mu^o  \right \|_2^2 }}
\end{equation}
Here, $\mu^m$ ($\sigma^2$) and $\tilde{\mu^m}$ ($\tilde{\sigma^2}$) are the mean (variance) representations of the implicit and inferential/implied form for $m^{th}$ implicit cluster, respectively. 

The above equation can be broken into two parts. The first part is equivalent to \ADD\, thus focusing on reducing the intra-cluster distance within the implicit class. The second part brings the implicit class closer to its implied meaning. Meanwhile, in the case of explicit or non-hate clusters, there is no mapping to the inferential/implied cluster, and \ADDINF\ in Equation \ref{eq:mag_imp_sample} reduces to \ADD\ in Equation \ref{eq:magnet_single}.

\textbf{Novel component: focal weight.} Both \ADDINF\ and \ADD\ assign uniform weight to all samples under consideration. In contrast, we have established that some instances are closer to the boundary of the imposter clusters and harder to classify (i.e., contribute more to the loss). Inspired by the concept of focal cross-entropy \citep{8237586}, we improve the \ADDINF\ objective by introducing \ADDINFFOC. Under \ADDINFFOC, the loss on each sample is multiplied by a factor called the \textit{focused term} $(1-p^{ADD^{inf}}(r_d^m))^\gamma$. $\gamma$, a hyperparameter, acts as a magnifier. The formulation assigns uniform weight as $\gamma \rightarrow 0$, reducing to \ADDINF. Analogously, the focal term is paying ``more attention" to specific data points. Even without inferential infusion, our novel focal term can be incorporated as \ADDFOC\ as enlisted in Equation \ref{eq:focal_magnet}.
\begin{equation}
\label{eq:focal_magnet}
\ell^{ADD^{*}}(\Theta) = \frac{1}{MD} \sum_{m=1}^{M} \sum_{d=1}^{D}\bigg[-(1-p^{ADD^{*}}(r_d^m))^\gamma log(p^{ADD^{*}}(r_d^m))\bigg]
\end{equation}

Here, \textcolor{black}{$\ell^{ADD^{inf+foc}}$ ($\ell^{ADD^{foc}}$) captures the setup with (without) inferential objective.} We utilise $p^{ADD^{inf}}$ (Equation \ref{eq:mag_imp_sample}) for the former and $p^{ADD}$ (Equation \ref{eq:magnet_single}) for the latter. Despite \ADDFOC\ being a minor update on \ADD, we empirically observe that focal infusion improves \ADD.

\textcolor{black}{\textbf{Training pipeline.}}
It should be noted that selecting the seed cluster and its subsequent imposter clusters is a random process for initial iterations. We assign the label with the highest loss margin for later iterations as the seed. Here, \ADDINFFOC operates for implicit hate and overcomes the drawback of existing literature where implicit detection fails to account for implied context. It is also essential to point out that this evaluation is carried out in the local neighborhood, which is aided by the focal loss (Equation \ref{eq:focal_magnet}).

\textbf{Overall loss.}
Apart from employing $r_d$ in $\ell^{ADD^{*}}$ it is also passed through a classification head $CH(r_d)$. We combine cross entropy with the focal inference to obtain the final loss of \model\ with $\beta$ controlling the contribution of the two losses as Equation \ref{eq:overall_loss}.

\begin{equation}
\label{eq:overall_loss}
\ell(\Theta)= \beta\ell^{CE}(\Theta) + (1-\beta)\ell^{ADD^{*}}(\Theta)
\end{equation}

\textbf{Inference.} During inference, the system does not have access to implied meaning. Once the PLM is trained via \model,  the classification head performs classification similar to any finetuned PLM. Here, we rely on the latent space being modified so that the implicit statements are closer to their semantic or implied form and sufficiently separated from other classes. 

\textcolor{black}{\textbf{Note on K-mean.}
As a clustering algorithm, K-means is the most generic as it does not assume any dataset property (like hierarchy) except for the semantic similarity of the samples. Further, the K-means computation happens offline in each epoch, i.e., it does not consume GPU resources. In the future, we aim to employ faster versions of K-means to improve training latency. Meanwhile, the computational complexity of \model\ during inference is the same as the finetuned PLM.}  

\section{Experimental Setup}
\label{sec:exp}
\model\ provides an opportunity to improve the detection of implicit context. In the first set of experiments, we focus on the task of hate speech classification with datasets that consist of implicit hate labels. In the second set of experiments, we establish the generalizability of the proposed framework via SemEval\footnote{\url{https://semeval.github.io/}} datasets on three separate tasks. Table \ref{tab:datasets_stats} provides both sets' label-wise distribution. In all the tasks, the surface form of the text varies contextually from its semantic structure. \textcolor{black}{Besides introducing the datasets and annotation schema, this section also outlines the hyperparameters and baselines curated for our evaluation.}

\textbf{Implicit hate classification datasets.} \textcolor{black}{Based on our literature survey of implicit hate datasets, we discard the ones that are either multilingual (DALC) or multimodal (ConvAbuse, MMHS150K), as modeling them is out of the scope of current work. Further, SBIC and ToxiGen do not offer 3-way labels; hence, they are discarded, too. From among the English datasets left for assessment,  we drop ISHate as it is an aggregated dataset, and its implicit samples are already covered by \latent. 
Finally, we have \latent\, \abuse, and \gab\ as English-based text-only datasets with explicit, implicit, and non-hate labels that suit our task.} For \latent, we employ the first level of annotation and the existing manual annotations of implied hatred for implicit samples. Meanwhile, \abuse\ and \gab\ do not have the implied descriptions. We manually annotate the implicit samples of these datasets with their implied meaning generated as free text.

\begin{table*}
\caption{Some sample posts from \abuse\ and \gab\ along with their implied annotations. We also provide the cross-annotator scores and the cross-annotator remarks.}
\label{tab:annot_quality}
\resizebox{\textwidth}{!}{
\begin{tabular}{|l|>{\arraybackslash}p{0.8\columnwidth}|>{\arraybackslash}p{0.3\columnwidth}|c|>{\arraybackslash}p{0.4\columnwidth}|}
\hline
\textbf{Dataset}&\textbf{Implicit Post}&\textbf{Implied Annotation}&\textbf{Score}&\textbf{Remarks}\\
\hline
\abuse & @USER How dare you tell me what Antifa is doing is right. They are not fighting fascism. They are fighting for their own stupid cause. They are further from fixing fascism than the Republicans themselves, & Anti-fascists are stupid and good for nothing. & \cellcolor{LightGreen}5 & Captured explicit target entity and the broader abuse in context \\ \cdashline{1-5}
\abuse &  @USER And this traitor thinks he is? & He thinks highly of himself. & \cellcolor{LightRed}2 &  Captured explicit target entity but fails to note broader abuse (traitor or anti-national) in context\\ \hline
\gab & I said it, and I stand by it! Blacks get almost all of their cultural cues and information from rap and sports. It's not like they're reading books or anything. URL & Black people don't read books. Black people are less intellectual &\cellcolor{LightGreen}5 & Target is identified correctly. An alternate implication can be: Black people are illiterate.\\ \cdashline{1-5}
\gab & You missed out on his post saying there are still many Jews on \#Gab! And didn't see his followers cuss out kike?! Good that you didn't see he claims himself a "racist capitalist"! I'm glad you missed ALL of them so you can continue to support him. AND excuse me, but he doesn't have much intelligence based on some of his posts. I'm not kike \& not stand w/ any sides! & Christian are anti-Jews  & \cellcolor{LightRed}3 & Could be referring to a specific right-winged politician. Alternative meaning can be: You are as racist and anti-semitic as the leaders you follow. \\ \hline
\end{tabular}
}
\end{table*}

\textbf{Annotation for implied hate.} Implied contexts are succinct statements that make explicit the underlying stereotype. Note that the implied context cannot be considered a comprehensive explanation for implicit hate but rather a more explicit understanding of the underlying subtle connotations. For \abuse\ and \gab, two expert annotators (\textcolor{black}{one male and one female social media expert}; age range between 29 to 35) perform the annotations based on the following guidelines:
\begin{itemize}[noitemsep, nolistsep, leftmargin=1em] 
    \item Implied meaning should consider the post's author's perspective.
    \item Implied meaning should emphasize \textcolor{black}{on} the post's content only.
    \item Annotations must be explicitly associated with the target entity.
    \item Annotations must contain a broader abusive context for the given post.
    \item Annotations should balance lexical diversity and uniformity w.r.t abuse towards a target group.
\end{itemize}

\textbf{Annotation agreement.} For our use case, annotation agreement scores help establish how well-aligned and coherent the explicit connotations are. To carry out the assessment, annotators A and B exchange a random sample of $30$ annotation pairs. They score the pairs on a 5-point Likert scale \citep{likert}, with 5 being the highest agreement. For \abuse, we obtain a mean agreement of $4.13 \pm 1.13$, and $4.07 \pm 1.41$ for \gab. Table \ref{tab:annot_quality} lists some sample annotations and their agreement scores. \textcolor{black}{Further, a third expert (a 24-year-old male) conducts an independent survey using the above metric on the other set of random $30$ samples. As per annotator C, for \abuse, we obtain a mean agreement of $4.55 \pm 1.09$ and $4.41 \pm 1.15$ for \gab. This independent assessment corroborates the annotation process, as annotator C  did not participate in the initial annotations yet observed similar alignment scores.}

\textbf{Generalizibility testing.} We further consider three SemEval tasks for our generalizability analysis. \textcolor{black}{Sarcasm detection \citep{abu-farha-etal-2022-SemEval}} and irony detection \citep{van-hee-etal-2018-SemEval} are two-way classification datasets. Meanwhile, stance detection \citep{mohammad-etal-2016-SemEval} is a three-way classification. While we have implied annotations for sarcasm, they are missing for the other two datasets. Here, no additional annotations are performed.

\textcolor{black}{\textbf{Hyperparameters.}} We run all experiments on two Nvidia V100 GPUs. Three random seeds (1, 4, 7) are used per setup. We report each setup's best performance based on overall macro-F1 out of three random seeds, where the best seed for a setup may vary. We follow an 80-20 split for the dataset across experiments (specific to the seed). \textcolor{black}{In initial experiments, we observe that \ADDINFFOC\ has a stronger influence on the later iterations, whereas CE influences the initial ones. Thus, to balance them throughout the training process, we put equal weightage on both using $\beta = 0.5$.} We consider $K=3$ with $M=2$ imposters for all experiments. We leave the experiments for $\beta$ and $M$ for future work. We set $100$ as the maximum K-means iterations in each training step. During finetuning, each training cycle is executed for a maximum of $5000$ epochs with all layers of  PLM frozen. 

\begin{table}
\centering
\caption{Dataset\textcolor{black}{s} employed in evaluating the \model\ framework. The statistics enlist the class-wise distributions for (a) Hate Speech and (b) SemEval datasets.}
\label{tab:datasets_stats}
\subfloat[]{
\resizebox{0.3\columnwidth}{!}{
\begin{tabular}{|l|c|ccc|}
\hline
{\textbf{Datasets}}
      & \textbf{\textcolor{black}{source}} & \multicolumn{3}{c|}{\textbf{Labels}} \\
  \hline   
& & N-OFF & EXP & IMP\\
\abuse\ & \textcolor{black}{Twitter} & 11173 &2129&798  \\
\cline{1-5}
& & N-OFF & EXP & IMP\\
\gab\ & \textcolor{black}{Gab} & 25102 &2159& 404 \\
\cline{1-5}
& & N-OFF & EXP & IMP\\
\latent\ & \textcolor{black}{Twitter} & 13291 &1089& 7100 \\
\hline
\end{tabular}}}
\quad
\quad
\subfloat[]{
\resizebox{0.3\columnwidth}{!}{
\begin{tabular}{|l|ccc|}
\hline
{\textbf{Task}}
      & \multicolumn{3}{c|}{\textbf{Labels}} \\
  \hline   
& N-SAR & SAR & \\
Sarcasm &3801 &1067&  \\
\cline{1-4}
& N-IRO & IRO &\\
Irony &1890 &1756&  \\
\cline{1-4}
& NEU & ANG & FAV\\
Stance &918 &1969& 982 \\
\hline
\end{tabular}}}
\vspace{-3mm}
\end{table}

\begin{table}
\centering
\caption{\textcolor{black}{Baseline selection based on comparison of $ADD^{foc}$ over: (a) vanilla \ADD\ for two-way hate speech classification via LSTM. (b) ACE for three-way hate speech classification via BERT.}}
\label{tab:baselines}
\subfloat[]{
\resizebox{0.425\columnwidth}{!}{
\begin{tabular}{|l|l|c|c|c|}
\hline
\multirow{3}{*}{\textbf{Dataset}} & \multirow{3}{*}{\textbf{Metric (F1)}}
      & \multicolumn{3}{c|}{\textbf{LSTM}} \\ \cline{3-5}
      
& & \ADD\ & $\alpha$-$ADD$ & $ADD^{foc}$\\ \hline
\multirow{3}{*}{\latent
} & Macro &0.557&0.601&\textbf{0.665}  \\
 & N-Hate &0.646&0.695&0.782  \\
 & Hate &0.469 & 0.509 &0.629  \\ \hline
\multirow{3}{*}{\textcolor{black}{\gab}} &
Macro &0.554&0.566&\textbf{0.639}  \\
 & N-Hate &0.919&0.911&0.927  \\
 & Hate &0.189&0.222&0.351  \\ \hline
\multirow{3}{*}{\abuse} & Macro &0.552&0.554&\textbf{0.634}  \\
 & N-Hate &0.820&0.780&0.851  \\
 & Hate &0.284&0.327& 0.417 \\ \hline
\end{tabular}}}
\subfloat[]{
\resizebox{0.425\columnwidth}{!}{
\begin{tabular}{|l|l|c|c|}
\hline
\textbf{Dataset}&\textbf{Metric (F1)}& $ADD^{foc}$ & $ACE$\\
 \hline
\multirow{3}{*}{\latent} & Macro &0.457&\textbf{0.533}  \\
 & N-Hate &0.690&0.772\\
 & EXP &0.221 & 0.268   \\
 & IMP &0.462 & 0.558  \\\hline
\multirow{3}{*}{\textcolor{black}{\gab}} &
Macro &0.451&\textbf{0.470}  \\
 & N-Hate &0.926&0.924  \\
& EXP &0.347 & 0.390  \\
 & IMP &0.080 & 0.097  \\\hline
\multirow{3}{*}{\abuse} & Macro &0.504&\textbf{0.530}  \\
 & N-Hate &0.817&0.854  \\
 & EXP &0.491 & 0.503   \\
 & IMP &0.203 & 0.233  \\\hline
\end{tabular}}}
\vspace{-3mm}
\end{table}

\textcolor{black}{\textbf{PLMs.}} \textcolor{black}{We begin our assessment with BERT \citep{devlin-etal-2019-bert}. For hate speech detection, we also employ a domain-specific HateBERT \citep{caselli-etal-2021-HateBERT} model to establish generalizability beyond BERT embedding.} HateBERT is built upon the concepts of continued pretraining on top of BERT. Here, the corpus for performing another round of unsupervised masking language modeling is obtained from potentially offensive subreddits. For the SemEval tasks, \textcolor{black}{we consider BERT and XLM \citep{chi-etal-2022-xlm} for evaluation based on their popularity in the SemEval.} The PLMs variants are `bert-base-uncased' for BERT, `xlm-roberta-large' for XLM, and `GroNLP/hateBERT' for HateBERT. 

\textbf{\textcolor{black}{Baselines.}} First \textcolor{black}{we assess} the improvement in \textcolor{black}{the performance} of $ADD^{foc}$ over vanilla \ADD\ (Equation \ref{eq:focal_magnet}) without the influence of cross-entropy. We follow the same prediction setup adopted in \ADD\ \citep{magnetloss}, where a sample gets assigned the label based on the nearest cluster in trained latent space during inference. We choose a simple LSTM \citep{Hochreiter1997} model for quicker experimentation and compare the original \ADD\ formulation with class weighted \ADD\ ($\alpha$-\ADD) and our proposed \ADDFOC. Table \ref{tab:baselines}(a) shows a significant performance improvement of $8.2$-$10.8$\% in overall macro-F1 using \ADDFOC\ across all three hate datasets. We thus recommend using our \ADDFOC\ variant instead of vanilla \ADD\ for future works. Interestingly, we note that $\alpha$-\ADD\ does not outperform \ADDFOC. Hence, it is not employed in further experiments. Further, we perform a 3-way classification using BERT to compare standalone ACE against standalone \ADDFOC. The results are presented in Table \ref{tab:baselines}(b). We observe that alpha cross-entropy (ACE) outperforms the standalone $ADD^{foc}$ by a substantial margin of $7.6\%$, $1.9\%$ and $2.6\%$ for \latent, \gab\ and \abuse\ respectively. Based on the above two experiments, we employ ACE as our baseline. As the proposed model introduces an additional loss complimenting ACE, we thus use ACE+$ADD^{foc}$ variants as comparative systems. 

\section{Results and Abaltions}
\label{sec:results}
In this section, we enlist the performance of \model\ for classifying implicit hate and discuss its robustness under different tasks and ablation setups. In both two and three-way hate classification, clustering is performed w.r.t. the three classes; however, the classification head is determined by the specific setup, either two or three-way. \textcolor{black}{For two-way hate classification, explicit (EXP) and implicit (IMP) labels are consolidated under the \textit{Hate} class.}

\textcolor{black}{\textbf{Two-way hate classification.}} From Table \ref{tab:two-way-cls}, we note that \model\ variants improves overall macro-F1 by $0.58$ ($\uparrow0.83$\%), $2.47$ ($\uparrow3.68$\%), and $0.56$ ($\uparrow0.79$\%) in \latent, \gab, and \abuse, respectively using BERT. However, except for maximizing hate macro-F1, the inferential objective does not significantly impact the final macro-F1 in the case of a two-way classification. It can be explained by the partially conflicting objectives between the final two-way result and \ADDINFFOC's three-way objective, leading to higher misclassification.  

\begin{table*}[!t]
\centering
\caption{\textcolor{black}{Results for two-way hate classification on BERT and HateBERT. We also highlight the highest Hate class macro-F1 that the respective model can achieve.}}
\label{tab:two-way-cls}
\resizebox{\columnwidth}{!}{
\begin{tabular}{|l|l|c|c|c|c|c|c|}
\hline

\multirow{3}{*}{\textbf{Dataset}} & \multirow{3}{*}{\textbf{Metric (F1)}}
      & \multicolumn{3}{c|}{\textbf{BERT}} &
      \multicolumn{3}{c|}{\textbf{HateBERT}} \\
       \cline{3-8}
 
& & \textbf{ACE} & $ \textbf{ACE+ADD\textsuperscript{{\tiny\textbf{foc}}}}$ & $ \textbf{ACE+ADD\textsuperscript{{\tiny\textbf{inf + foc}}}} $ & \textbf{ACE} &  $\textbf{ACE+ADD\textsuperscript{{\tiny\textbf{foc}}}} $ & $ \textbf{ACE+ADD\textsuperscript{{\tiny\textbf{inf + foc}}}} $ \\
\hline
\multirow{4}{*}{\latent} & Macro &0.6991    &\textbf{0.7049}    &0.7039 &0.7121 &\textbf{0.7135}    &0.7121  \\
& N-Hate &0.7599    &0.7603 &0.7638 &0.7843 &0.7787 &0.7791  \\
& Hate &0.6383  &\textbf{0.6495}    &0.6439 &0.6400 &\textbf{0.6484}    &0.6450 \\ \cdashline{2-8}
& Highest Hate F1 &0.6478   &0.6491 &\textbf{0.6548}    &0.6615 &0.6572 &\textbf{0.6630}  \\\hline   
\multirow{4}{*}{\gab} & 
Macro &0.6709   &0.6886 &\textbf{0.6956}    &0.6889 &\textbf{0.7027}    &0.6987 \\
& N-Hate &0.9171    &0.9313 &0.9355 &0.9239 &0.9398 &0.9344  \\
& Hate &0.4247  &0.4459 &\textbf{0.4556}    &0.4538 &\textbf{0.4657}    &0.4631 \\\cdashline{2-8}
& Highest Hate F1 &0.4281   &0.4494 &\textbf{0.4562}    &0.4538 &0.4663 &\textbf{0.4631} \\
   \hline

\multirow{4}{*}{\abuse} & Macro &0.7075 &\textbf{0.7131}    &0.7124 &0.7189 &0.7176 &\textbf{0.7202}  \\
 & N-Hate &0.8729   &0.8779 &0.8675 &0.8767 &0.8711 &0.8808  \\
 & Hate & 0.5420    &0.5483 &\textbf{0.5574}    &0.5610 &\textbf{0.5641}    &0.5595 \\\cdashline{2-8}
   & Highest Hate F1 &0.5502    &0.5583 &\textbf{0.5625}    &0.5642 &\textbf{0.5710}    &0.5691 \\
\hline
\end{tabular}}
\vspace{-2mm}
\end{table*}

\begin{table*}[!t]
\centering
\caption{
\textcolor{black}{Results for three-way hate classification on BERT and HateBERT.
}}
\label{tab:three-way-cls}
\resizebox{\columnwidth}{!}{
\begin{tabular}{|l|l|c|c|c|c|c|c|}
\hline

\multirow{3}{*}{\textbf{Dataset}} & \multirow{3}{*}{\textbf{Metric (F1)}}
      & \multicolumn{3}{c|}{\textbf{BERT}} &
      \multicolumn{3}{c|}{\textbf{HateBERT}} \\
       \cline{3-8}
 
& & \textbf{ACE} & $ \textbf{ACE+ADD\textsuperscript{{\tiny\textbf{foc}}}}$ & $ \textbf{ACE+ADD\textsuperscript{{\tiny\textbf{inf + foc}}}} $ & \textbf{ACE} &  $\textbf{ACE+ADD\textsuperscript{{\tiny\textbf{foc}}}} $ & $ \textbf{ACE+ADD\textsuperscript{{\tiny\textbf{inf + foc}}}} $ \\

\hline
\multirow{5}{*}{\latent} & Macro &0.5327    &0.5331 &\textbf{0.5336}    &0.5537 &\textbf{0.5607}    &0.5571  \\
 & N-Hate &0.7722   &0.7609 &0.7474 &0.7480 &0.7654 &0.7829 \\
 & EXP &0.2682  &0.2775 &0.2776 &0.3403 &0.3300 &0.3111 \\
  & IMP &0.5577 &0.5610 &\textbf{0.5759}    &0.5728 &\textbf{0.5866}    &0.5774 \\
   \hline

\multirow{5}{*}{\gab} & Macro 
&0.4621 &0.4575 &\textbf{0.4668}    &0.4797 &0.4772 &\textbf{0.4813} \\
 & N-Hate &0.9198   &0.9197 &0.9258 &0.9230 &0.9434 &0.9339  \\
 & EXP &0.3775  &0.3850 &0.3819 &0.4055 &0.4076 &0.4190 \\
  & IMP &0.0889 &0.0678 &\textbf{0.0928}    &\textbf{0.1105}    &0.0806 &0.0909 \\ \hline
   \multirow{5}{*}{\abuse} & Macro &0.5300  &0.5328 &\textbf{0.5398}    &0.5309 &0.5311&    \textbf{0.5313}  \\
 & N-Hate &0.8541   &0.8370 &0.8867 &0.8532 &0.8611 &0.8606  \\
 & EXP &0.5027  &0.5256 &0.5234 &0.5038 &0.5138 &0.5111 \\
  & IMP & 0.2333    & \textbf{0.2359}   &0.2094 &\textbf{0.2357}    &0.2185 &0.2222  \\
\hline
\end{tabular}}
\end{table*}

\textcolor{black}{\textbf{Three-way hate classification.}} Inferential infusion reasonably impacts the outcome of the three-way classification task (Table \ref{tab:three-way-cls}). 
Overall, in 3-way classification, \ADDINFFOC\ provide an improvement of $0.09$ ($\uparrow0.17$\%), $0.47$ ($\uparrow1.02$\%), and $0.98$ ($\uparrow1.85$\%) in macro-F1 for \latent, \gab, and \abuse, respectively, on BERT. \emph{It is noteworthy that we observe an even higher level of improvement for implicit hate class than overall.} Compared to ACE in three-way classification, \ADDFOC\ helps \abuse\ with an improvement of $0.26$ macro-F1 ($\uparrow1.11$\%) in implicit hate. Meanwhile, \ADDINFFOC\ helps \latent\ and \gab\ with an improvement of $1.82$ ($\uparrow3.26$\%) and $0.39$ ($\uparrow4.39$\%), macro-F1 respectively, in implicit hate. 

\textbf{Generalizability Test.}
The availability of implied annotations in the sarcasm dataset enables us to test \model's $ADD^{Inf+foc}$ variant. The unavailability of such annotations in the other two tasks limits our experiments with only $ADD^{foc}$ variant. Table \ref{tab:res_sarcasm} (a) and (b) present the results for sarcasm detection and the other two tasks (irony and stance detection), respectively. Barring one setup, we observe reasonable improvements in macro-F1 (0.41-2.37) across all three tasks using both PLMs. Further, for the minority class considering the best of the BERT and XLM, we observe \model\ variants report an improvement of $6.06$ ($\uparrow23.96$)\%, $1.35$ ($\uparrow2.65$\%) and $3.14$ ($\uparrow5.42$\%) for the respective minority class in sarcasm, stance and irony detection.

\begin{table} [!h]
\caption{Comparative performance for sarcasm, irony, and stance detection.}
\label{tab:res_sarcasm}
\subfloat[]{
\resizebox{0.45\columnwidth}{!}{
\begin{tabular}{|l|l|l|c|c|}
\hline
\multirow{3}{*}{\textbf{Tasks}}& \multirow{3}{*}{\textbf{Model}} & \multirow{3}{*}{\textbf{Metric (F1)}}
 & \multicolumn{2}{c|}{\textbf{Objective}}\\ \cline{4-5}
& & & \textbf{ACE} & $ \textbf{ACE+ADD\textsuperscript{{\tiny\textbf{inf + foc}}}} $ \\ \hline
\multirow{6}{*}{Sarcasm} &\multirow{3}{*}{BERT} 
&Macro &\textbf{0.5651}&0.5610 \\
 && N-SAR &0.8329&0.8501 \\
 && SAR &0.2974&0.2720  \\ \cline{3-5}
 &\multirow{3}{*}{XLM} 
&Macro &0.5577&\textbf{0.5814}  \\
 && N-SAR &0.8626&0.8493  \\
 && SAR &0.2529&0.3135  \\
\hline
\end{tabular}}}
\quad
\subfloat[]{
\resizebox{0.4\columnwidth}{!}{
\begin{tabular}{|l|l|l|c|c|}
\hline
\multirow{3}{*}{\textbf{Tasks}}& \multirow{3}{*}{\textbf{Model}} & \multirow{3}{*}{\textbf{Metric (F1)}}
 & \multicolumn{2}{c|}{\textbf{Objective}}\\ \cline{4-5}

& & & \textbf{ACE} & $ \textbf{ACE+ADD\textsuperscript{{\tiny\textbf{foc}}}}$ \\ \hline
   \multirow{8}{*}{Stance} &\multirow{4}{*}{BERT} &Macro &0.5735&\textbf{0.5776} \\
 && NEU &0.5009&0.4910 \\
 && ANG &0.6964&0.7090  \\
   && FAV &0.5231&0.5327 \\ \cline{3-5}
   &\multirow{4}{*}{XLM} &Macro &0.5560&\textbf{0.5692}  \\
 && NEU &0.4494&0.4637  \\
 && ANG &0.7084&0.7199  \\
   && FAV &0.5103&0.5238  \\  \hline
\multirow{6}{*}{Irony} &\multirow{3}{*}{BERT} 
&Macro &0.6886&\textbf{0.7099}  \\
 && N-IRO &0.7379&0.7658 \\
 && IRO &0.6394&0.6540  \\ \cline{3-5}
 &\multirow{3}{*}{XLM} 
&Macro &0.6637&\textbf{0.6740}  \\
 && N-IRO &0.7482&0.7372  \\
 && IRO &0.5792&0.6106  \\ \hline
\end{tabular}}}
\vspace{-3mm}
\end{table}

\textbf{\textcolor{black}{Impact of domain-specific PLM.}}
Under HateBERT, \model\ variants improve two-way classification by overall $0.14$ ($\uparrow0.20$\%), $1.38$ ($\uparrow2.00$\%) and $0.13$ ($\uparrow0.18$\%) for \latent, \gab, and \abuse\ respectively. Similarly, \model\ variants improve three-way classification by overall $0.7$ ($\uparrow1.26$\%), $0.16$ ($\uparrow0.34$\%), $0.04$ ($\uparrow0.08$\%) for \latent, \gab, and \abuse\ respectively. However, the results with HateBERT show more variability. While all datasets in two-classification via HateBERT benefit from \model\, implicitness of \abuse, and \gab\ suffer under three-way classification. This variation can be attributed to a lot more offensive and slur terms in HateBERT's training than BERT. Through this analysis, we are able to comment on the domain-specific (HateBERT) vs general-purpose (BERT) systems and their role in finetuning. Interestingly, this has been noted in other research in hate speech as well \citep{masud-etal-2024-probing}.

On the other hand, under generalization testing, which utilizes only general-purpose encoders (BERT and XLM), a high-performance improvement is observed in all minority classes. 

\begin{figure}[!t]
\centering
\subfloat[]{\includegraphics[width=0.375\textwidth]{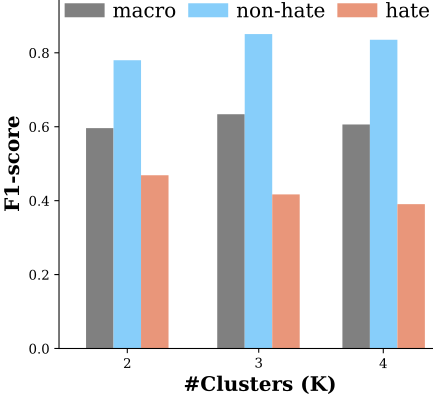}}
\hspace{1mm} 
\subfloat[]{\includegraphics[width=0.525\textwidth]{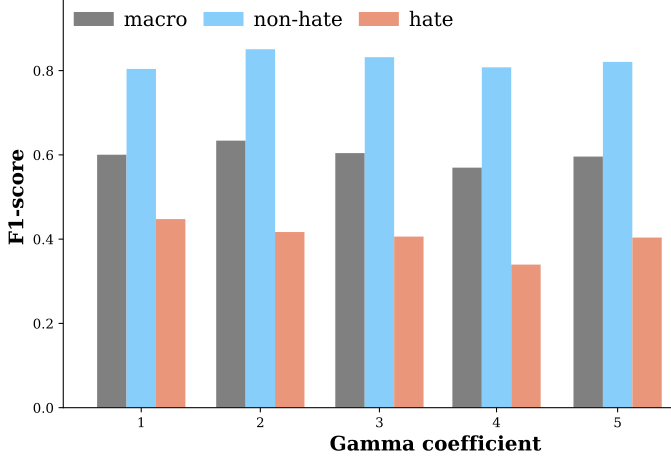}}
\caption{The variation in performance with changing values of (a) number of clusters (k) and (b) focal parameter ($\gamma$ ). We employ BERT on \abuse\ with $ADD^{foc}$ in the two-way classification.}
\label{fig:numk_gamma_coef}
\end{figure}

\begin{table*}[!h]
\tiny
\centering
\caption{\textcolor{black}{Results for two-way classification task across all three hate-speech datasets using two pretrained language models, BERT and HateBERT. Highlighted with green color are the outcomes where one of the variants of \model\ outperforms baseline ACE.}}
\label{tab:two-way-cls-detail}
\resizebox{\columnwidth}{!}{
\begin{tabular}{|l|l|l|c|c|c|c|c|c|}
\hline
\multirow{3}{*}{\textbf{Dataset}} & \multirow{3}{*}{\textbf{Seed}}&\multirow{3}{*}{\textbf{Metric (F1)}}
      & \multicolumn{3}{|c|}{\textbf{BERT}} &
      \multicolumn{3}{|c|}{\textbf{HateBERT}} \\ \cline{4-9}
 
& & & \textbf{ACE} & $ \textbf{ACE+ADD^{\tiny\textbf{foc}}}$ & $ \textbf{ACE+ADD^{\tiny\textbf{inf + foc}}} $ & \textbf{ACE} &  $\textbf{ACE+ADD^{\tiny\textbf{foc}}} $ & $ \textbf{ACE+ADD^{\tiny\textbf{inf + foc}}} $ \\ \hline
\multirow{9}{*}{\latent} & \multirow{3}{*}{1}& Macro &\cellcolor{LightGreen}0.69658&   \cellcolor{LightGreen}0.70291   &\cellcolor{LightGreen}0.69780  &\cellcolor{LightRed}0.70843&   \cellcolor{LightRed}0.69983 &
\cellcolor{LightRed}0.70143  \\
&& N-Hate &0.76353& 0.76010 &0.76069&   0.77514&0.75034 &0.76016  \\
&& Hate &0.62962&   0.64571&    0.63492&    0.64171&    0.64931&0.64271 \\ \cdashline{2-9}   
&\multirow{3}{*}{4} & 
Macro &\cellcolor{LightGreen}0.69388    &\cellcolor{LightGreen}0.68951  &\cellcolor{LightGreen}0.69611& \cellcolor{LightRed}0.70720&
\cellcolor{LightRed}0.70423&    \cellcolor{LightRed}0.70709\\
&& N-Hate &0.75595  &0.75246    &0.77109    &0.76370&   0.75511 &0.76389  \\
&& Hate &0.63182    &0.62656    &0.62113    &0.65070&   0.65335 &0.65030 \\ \cdashline{2-9}
  &\multirow{3}{*}{7} & Macro &\cellcolor{LightGreen}0.69907    &\cellcolor{LightGreen}0.70490& \cellcolor{LightGreen}0.70387   &\cellcolor{LightGreen}0.71212& \cellcolor{LightGreen}0.71353   &\cellcolor{LightGreen}0.71205 \\
& & N-Hate &0.75987&    0.76033 &0.76384    &0.78429    &0.77868    &0.77907 \\
& & Hate & 0.63827  &0.64946    &0.64389    &0.63995&   0.64838&    0.64502 \\
\hline

\multirow{9}{*}{\gab} & \multirow{3}{*}{1}& Macro &\cellcolor{LightGreen}0.67087 &\cellcolor{LightGreen}0.68862  &\cellcolor{LightGreen}0.69557  &\cellcolor{LightGreen}0.68886  &\cellcolor{LightGreen}0.69790  &\cellcolor{LightGreen}0.69872  \\
&& N-Hate &0.91708  &0.93134    &0.93551    &0.92391    &0.92951    &0.93438 \\
&& Hate &0.42467    &0.44590    &0.45563    &0.45380    &0.46630    &0.46307\\ \cdashline{2-9} 

&\multirow{3}{*}{4} & 
Macro &\cellcolor{LightGreen}0.66411    &\cellcolor{LightGreen}0.67941  &\cellcolor{LightGreen}0.67591  &\cellcolor{LightGreen}0.68248  &\cellcolor{LightGreen}0.67406  &\cellcolor{LightGreen}0.68307\\
&& N-Hate &0.91361  &0.92403    &0.91952    &0.92069    &0.91105&   0.92318 \\
&& Hate &0.41461    &0.43478&   0.43231&    0.44428 &0.43708&   0.44295\\ \cdashline{2-9}   

  &\multirow{3}{*}{7} & Macro &\cellcolor{LightGreen}0.66402    &\cellcolor{LightGreen}0.68237& \cellcolor{LightGreen}0.68206   &\cellcolor{LightGreen}0.68720& \cellcolor{LightGreen}0.70271&  \cellcolor{LightGreen}0.68962  \\
& & N-Hate &0.91019&    0.92933&    0.92782&    0.92266&    0.93976&    0.93049  \\
& & Hate & 0.41784& 0.43541&    0.43630&    0.45175&    0.46565 &0.44874 \\
\hline

\multirow{9}{*}{\abuse} & \multirow{3}{*}{1}& Macro &\cellcolor{LightGreen}0.69253   &\cellcolor{LightGreen}0.70627  &\cellcolor{LightGreen}0.70502  &\cellcolor{LightGreen}0.70633  &\cellcolor{LightGreen}0.71030& \cellcolor{LightGreen}0.70699 \\
&& N-Hate &0.85193  &0.86709    &0.87395    &0.87774    &0.87652&   0.88362  \\
&& Hate &0.53313&   0.54545 &0.53609    &0.53492&   0.54409&    0.53035\\ \cdashline{2-9}  
&\multirow{3}{*}{4} & 
Macro &\cellcolor{LightGreen}0.70748&   \cellcolor{LightGreen}0.71313   &\cellcolor{LightGreen}0.71244  &   \cellcolor{LightGreen}0.71885   &   \cellcolor{LightGreen}0.71761   &   \cellcolor{LightGreen}0.72018\\
&& N-Hate &0.87293& 0.87792 &0.86754    &   0.87667 &0.87112        &0.88082 \\
&& Hate &0.54204&   0.54833 &0.55735    &   0.56103     &0.56410        &0.55953\\ \cdashline{2-9}   

  &\multirow{3}{*}{7} & Macro &\cellcolor{LightGreen}0.67794    &   \cellcolor{LightGreen}0.68587   &   \cellcolor{LightGreen}0.68944   &   \cellcolor{LightGreen}0.69350   &   \cellcolor{LightGreen}0.70146&      \cellcolor{LightGreen} 0.69910  \\
& & N-Hate &0.85393     &0.86020    &   0.85694 &   0.85240 &   0.85052 &0.86129  \\
& & Hate & 0.67794  &   0.68587 &   0.68944 &   0.69350 &   0.70146     &0.69910 \\
\hline
\end{tabular}}
\end{table*}

\textbf{Significance of hyperparameters.}
We further experiment with the hyperparameters of the \model. The experiments are performed on a two-way hate classification task on the \abuse\ dataset using BERT. The limited range for the probe is heuristically defined based on the sample size of categories. We recommend determining the values on a case-to-case basis for optimized performance. Figure \ref{fig:numk_gamma_coef}(a) represents the significance of the number of subclusters per class ($k$) in the range of [2-4]. We observe comparable performance for $K=3$ or $4$. For our experiments, \textcolor{black}{since four} of the six datasets contain three classes, we use $K=3$. The intuition is that within a subcluster of a class, the three subclusters represent a case of one of them having a high affinity to the class itself and two others being closer to their imposter classes. For example, within the implicit hate class, we assume at least one subcluster is easy to label as implicit, while there will likely be at least one cluster each that is closer to explicit and non-hate classes. Consequently, the setup leads to an imposter cluster value of $M=2$. Meanwhile, the significance of the $\gamma$ coefficient used in the focused objective is presented in Figure \ref{fig:numk_gamma_coef}(b). The probe is limited to [1-5] with a unit interval \textcolor{black}{as followed in existing literature \citep{8237586}.} We observe the best outcome with $\gamma=2$ which incidentally aligns with the best value identified by  \citep{8237586}. 

\begin{table*}[!h]
\tiny
\centering
\caption{\textcolor{black}{Results for three-way classification tasks across all three hate-speech datasets using two pretrained language models, BERT and HateBERT. Highlighted with green color are the outcomes where one of the variants of \model\ outperforms baseline ACE.
}}
\label{tab:three-way-cls-detail}
\resizebox{\columnwidth}{!}{
\begin{tabular}{|l|l|l|c|c|c|c|c|c|}
\hline

\multirow{3}{*}{\textbf{Dataset}}&\multirow{3}{*}{\textbf{Seed}} & \multirow{3}{*}{\textbf{Metric (F1)}}
      & \multicolumn{3}{c}{\textbf{BERT}} &
      \multicolumn{3}{|c|}{\textbf{HateBERT}} \\
       \cline{4-9}
 
& & & \textbf{ACE} & $ \textbf{ACE+ADD^{\tiny\textbf{foc}}}$ & $ \textbf{ACE+ADD^{\tiny\textbf{inf + foc}}} $ & \textbf{ACE} &  $\textbf{ACE+ADD^{\tiny\textbf{foc}}} $ & $ \textbf{ACE+ADD^{\tiny\textbf{inf + foc}}} $ \\

\cline{1-9}
\multirow{12}{*}{\latent}&\multirow{5}{*}{1} & Macro &\cellcolor{LightGreen}0.50683    &\cellcolor{LightGreen}0.51872  &\cellcolor{LightGreen}0.51727  &\cellcolor{LightRed}0.53277    &\cellcolor{LightRed} 0.53023 & 0\cellcolor{LightRed}.52951  \\
 && N-Hate &0.76214 &0.76201    &0.74872    &0.76216    & 0.75547 & 0.76122 \\
 && EXP &0.20159    &0.25344    &0.25296    &0.25575 &  0.26912 &   0.26109 \\
  && IMP &0.55678   &0.54072    &0.55013    & 0.58041 & 0.56609 & 0.56622 \\ \cdashline{2-9}

&\multirow{5}{*}{4} & Macro 
&\cellcolor{LightGreen}0.52340  &\cellcolor{LightGreen}0.52746  &\cellcolor{LightGreen}0.52819  &\cellcolor{LightGreen}0.55372  & \cellcolor{LightGreen}0.56065 &   \cellcolor{LightGreen}0.54937 \\
 && N-Hate &0.77192 &0.75288    &0.76089    &0.74803    &0.76535&   0.75695  \\
 && EXP &0.26631    &0.25862    &0.28870    &0.34032    &0.33000&   0.32524 \\
  && IMP &0.53196   &0.57088    &0.53499    &0.57281    &0.58660&   0.56591 \\ \cdashline{2-9}

&  \multirow{5}{*}{7} & Macro &\cellcolor{LightGreen}0.53267    &\cellcolor{LightGreen}0.53312  &\cellcolor{LightGreen}0.53361  &\cellcolor{LightGreen}0.54265  &\cellcolor{LightGreen}0.54197& \cellcolor{LightGreen}0.55713  \\
 && N-Hate &0.77215 &0.76094    &0.74736    &0.76769&   0.77002 &0.78290  \\
 && EXP &0.26815    &0.27748    &0.27762    &0.29891&   0.28428&0.31111 \\
  && IMP & 0.55772  & 0.56095   &0.57587    &0.56136&   0.57162 &0.57738  \\ \hline

\multirow{12}{*}{\gab}&\multirow{5}{*}{1} & Macro &\cellcolor{LightGreen}0.46284&    \cellcolor{LightGreen}0.45740   &\cellcolor{LightGreen}0.47635  &\cellcolor{LightRed}0.47448    &\cellcolor{LightRed}0.47417    &\cellcolor{LightRed}0.47325  \\
 && N-Hate &0.90876 &0.91715&0.92309&0.92432    &0.92950&0.91882 \\
 && EXP &0.36140&   0.38783 &0.37983    &0.39767    &0.40897    &0.40336 \\
  && IMP &0.11834   &0.06722    &0.12612    &0.10144    &0.08403    &0.09756 \\ \cdashline{2-9}
  
&\multirow{5}{*}{4} & Macro 
&\cellcolor{LightGreen}0.45289  &\cellcolor{LightGreen}0.45772& \cellcolor{LightGreen}0.43606   &\cellcolor{LightGreen}0.46788  &\cellcolor{LightGreen}0.47415  &\cellcolor{LightGreen}0.44959 \\
 && N-Hate &0.92661&    0.91054&    0.91243 &0.92351    &0.92469&0.91982  \\
 && EXP &0.38078&   0.36646&    0.37195&0.38563 &0.42696    &0.40542 \\
  && IMP &0.05128&  0.09615 &0.02380    &0.09448&   0.07079 &0.02352 \\ \cdashline{2-9}

&  \multirow{5}{*}{7} & Macro &\cellcolor{LightGreen}0.47048    &\cellcolor{LightGreen}0.45741  &\cellcolor{LightGreen}0.46351  &\cellcolor{LightGreen}0.47969  &\cellcolor{LightGreen}0.47721&\cellcolor{LightGreen}0.48128  \\
 && N-Hate &0.92416 &0.93149&   0.93998 &0.92302&   0.94337 &0.93388  \\
 && EXP &0.39033&   0.40073 &0.38159    &0.40554&   0.40761&0.41904 \\
  && IMP & 0.09696  &0.04000&   0.06896&0.11049 &0.08064&   0.09090  \\ \hline

\multirow{12}{*}{\abuse}&\multirow{5}{*}{1} & Macro &\cellcolor{LightGreen}0.51761   &\cellcolor{LightGreen}0.52082  &\cellcolor{LightGreen}0.52249& \cellcolor{LightGreen}0.52462&  \cellcolor{LightGreen}0.52638&  \cellcolor{LightGreen}0.52559 \\
 && N-Hate &0.84885 &0.84862&   0.849518&   0.86772&    0.87497&    0.87748 \\
 && EXP &0.49009&   0.51483&    0.51033&    0.53555 &0.52680    &0.52727 \\
  && IMP &0.21390&  0.19900&    0.20765 &0.17058    &0.17737    &0.17204 \\ \cdashline{2-9}
&\multirow{5}{*}{4} & Macro 
&\cellcolor{LightGreen}0.53002& \cellcolor{LightGreen}0.53283&  \cellcolor{LightGreen}0.53982&  \cellcolor{LightGreen}0.52609&  \cellcolor{LightGreen}0.52095&  \cellcolor{LightGreen}0.52764\\
 && N-Hate &0.85409&    0.83699&    0.88670 &0.86536    &0.86110&   0.87867 \\
 && EXP &0.50272    &0.52564    &0.52338&   0.52133 &0.50485    &0.51759 \\
  && IMP &0.23325   &0.23587    &0.20938    &0.19158&   0.19689 &0.18666 \\ \cdashline{2-9}

&  \multirow{5}{*}{7} & Macro &\cellcolor{LightGreen}0.51914&   \cellcolor{LightGreen}0.52951&  \cellcolor{LightGreen}0.52534&  \cellcolor{LightGreen}0.53091 &\cellcolor{LightGreen}0.53110    &\cellcolor{LightGreen}0.53129  \\
 && N-Hate &0.84480&    0.85799&    0.85942 &0.85322&   0.86106&    0.86060 \\
 && EXP &0.50420&   0.51487&    0.51041&    0.50383 &0.51378    &0.51106\\
  && IMP & 0.20843& 0.21568&    0.20618&    0.23569 &0.21848    &0.22222  \\ \hline
\end{tabular}}
\vspace{-5mm}
\end{table*} 

\section{Does \model\ really improve implicit hate detection?}
\label{sec:analysis}
Given the overall macro-F1 results on hate speech detection vary in a narrow range, significance testing will be inconclusive. We thus perform a granular analysis of the results across all seeds and asses how well \model\ modifies the latent space. We also conduct an error analysis of cases where implicit hate is easy and hard to classify.

\textbf{Seed-wise Analysis.} \textcolor{black}{Across three random seeds, two PLMs, and three datasets, we record the performance for 18 setups, each in two-way and three-way hate speech detection. We note from Tables \ref{tab:two-way-cls-detail}  and \ref{tab:three-way-cls-detail} that out of the 36 combinations, only four instances register a drop in performance. It corroborates that \model's improvements are not limited to a specific initialization setup. Interestingly, the setups that register failure are all under HateBERT. The results further contribute to the discussion on domain-specific PLMs in Section \ref{sec:results}.}

\begin{figure*}[!h]
 \captionsetup[subfigure]{justification=centering}
    \centering
    \subfloat[Positive case]{\includegraphics[width=0.475\textwidth]{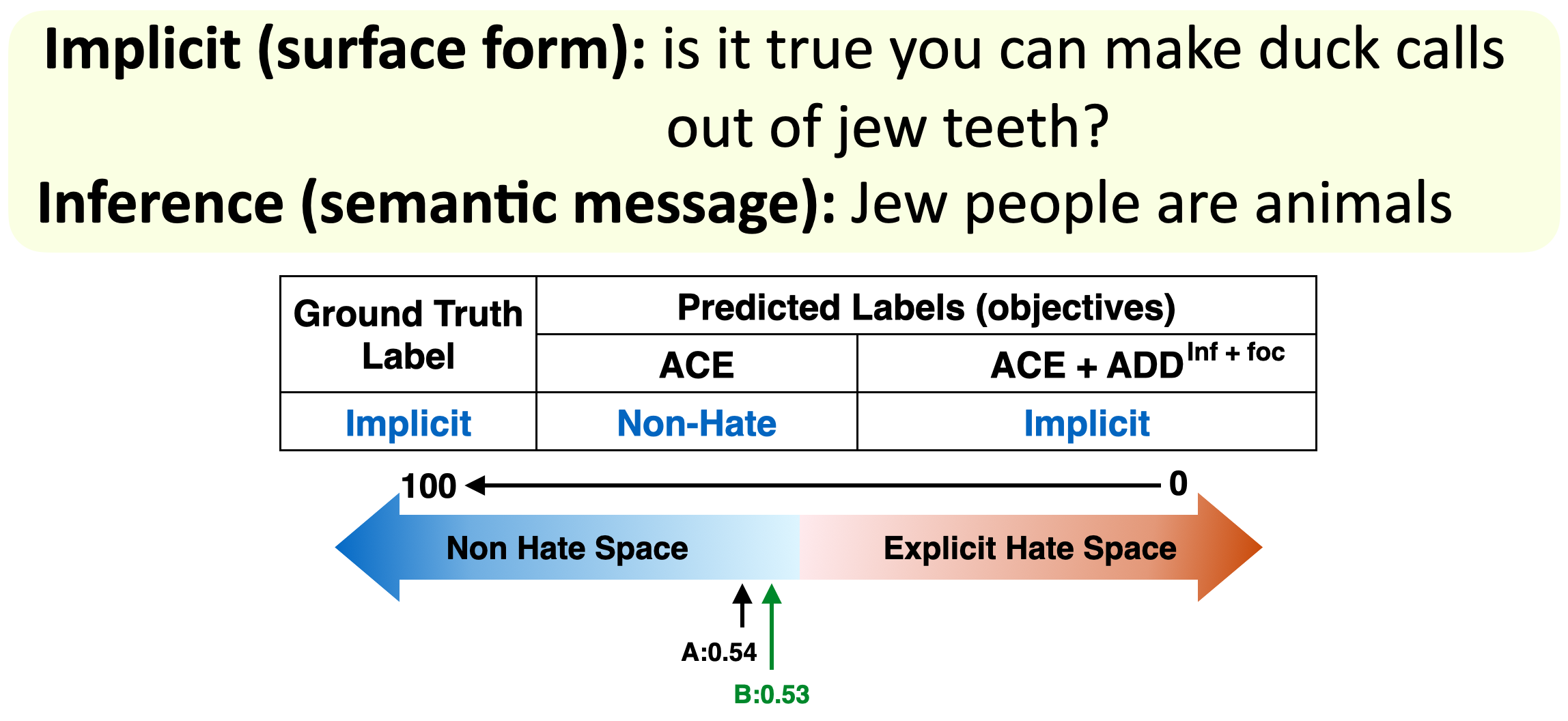}}
    \hspace{2mm} 
    \subfloat[Negative case]{\includegraphics[width=0.475\textwidth]{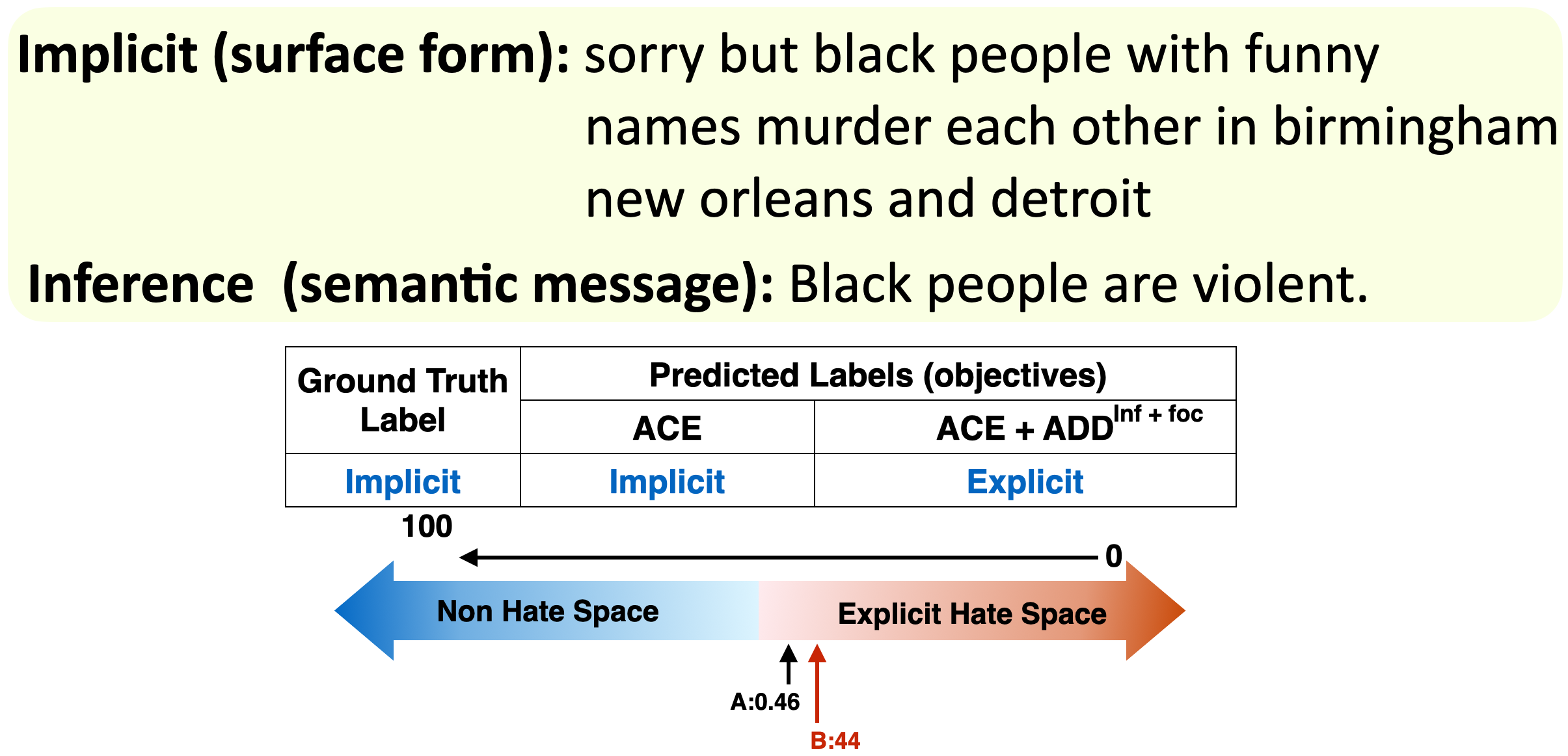}} 
    \caption{Error analysis with (a) correctly and (b) incorrectly classified samples in three-way classification on \latent. Here, scores A and B are the relative positions of implicit sample w.r.t non-hate and explicit space finetuned with ACE and \ADDINFFOC, respectively.}
    \label{fig:error_analysis}
\vspace{-3mm}
\end{figure*}

\textbf{Error Analysis.} The motivation for \model\ is that implicit is closer to non-hate than explicit hate. Employing \model\ should correct the misclassified implicit labels if this hypothesis holds. On the other hand, a false-positive may occur if the example is already close to explicit subspace. Further, moving it toward explicit space can cause misclassification. We, thus, consider a positive/negative case where the predicted label for an implicit sample is correctly/incorrectly classified. To explain these two scenarios, we estimate the relative distance of the implicit sample from explicit and non-hate clusters. First, we perform K-means clustering on non-hate and explicit latent space to identify their centers. We then calculate the average Manhattan distance between the implicit samples and these local density centers. Finally, we obtain the relative score from explicit space by normalizing between 0-1 the average explicit distance by the sum of average distances from non-hate and explicit spaces. For example, if the sample has a distance of 3 from explicit and 6 from non-hate centers, then the normalised distance will be $3/(3+6)=0.33$. 

We highlight a positive and a negative case in Figures \ref{fig:error_analysis} (a) and (b), respectively. In the positive case, the implicit sample is closer to non-hate space (Point A) under the ACE objective. After employing the \model, its relative position moves away from non-hate and closer to explicit (point B). In contrast, for the negative case, where the implicit sample is initially close to explicit hate (point A), our objective leads to misclassification. In the future, this problem can be reduced by introducing a constraint on the distance between implicit and explicit intact.

\subsection{Latent Space Analysis}
\label{sec:latent}
Building upon the cluster assessment in the error analysis, where we examined only a single positive and negative sample, we now perform an overall evaluation of how \ADDINFFOC\ manipulates the embedding space. Inspired by the existing literature examining the latent space under hate speech datasets \citep{fortuna-etal-2020-toxic} and models \citep{kim-etal-2022-generalizable, ocampo-etal-2023-depth}, we attempt to quantify the inter-cluster separation via Silhouette scores. 

\textbf{Silhouette score.} \textcolor{black}{It is a metric to measure the `goodness' of the clustering technique. It is calculated as a tradeoff between within-cluster similarity and inter-cluster dissimilarity.} Consider a system with E points ($e_i$), each point belonging to one of the N clusters $c^j(e_i)$. For $e_i \in c^a$, its Silhouette score $SS_i=\frac{max(p_i,q_i)}{q_i-p_i}$. $p_i$  captures the intra-cluster distance of $e_i$ to all the points within the cluster it belongs; $p_i = \frac{1}{|c^a|-1}\sum_{e^j \in c^a}dist(e_i,e_j)$. $q_i$ captures the inter-cluster distance of $e_i \in c^a$ to all the points in the nearest cluster to $c^a$; $q_i = \frac{1}{|c^b|}\sum_{e^j \in c^b}dist(e_i,e_j)$. The Silhouette score of a setup is, thus, $SS=\frac{1}{|E|}\sum_{e_i \in E}SS_i$.  Silhouette scores are measured on a scale of -1 to 1, with -1 being the worst set of cluster assignments.

\begin{figure*}[!h]
 \captionsetup[subfigure]{justification=centering}
    \centering
    \subfloat[Default BERT]{\includegraphics[width=0.356\textwidth]{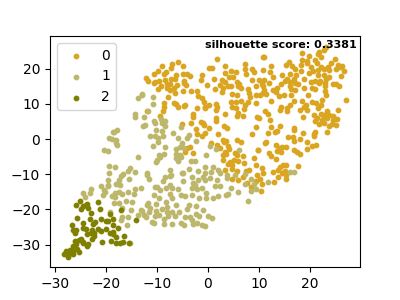}}
    \hspace{-5.75mm}
    \subfloat[Finetuned via $ACE$]{\includegraphics[width=0.356\textwidth]{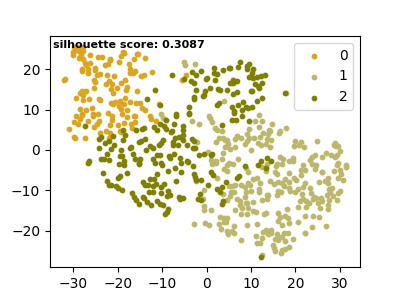}}
    \hspace{-5.75mm}
    \subfloat[Finetuned via $ACE + ADD^{Inf+foc}$]{\includegraphics[width=0.356\textwidth]{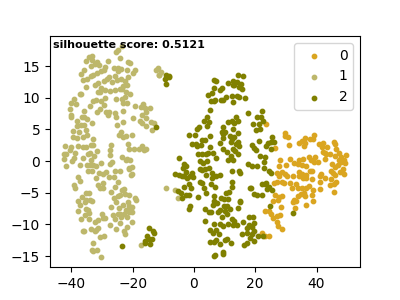}}\vspace{-4mm}
    
    \subfloat[Default BERT]{\includegraphics[width=0.356\textwidth]{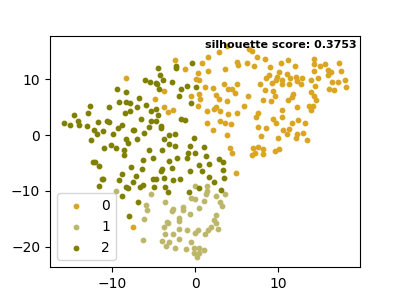}}\hspace{-5.75mm}
    \subfloat[Finetuned via ACE]{\includegraphics[width=0.356\textwidth]{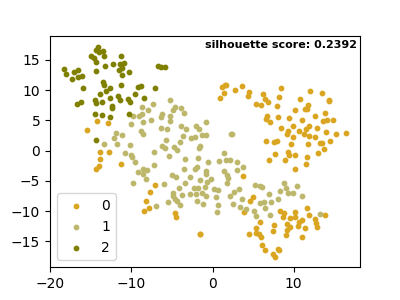}}\hspace{-5.75mm}
    \subfloat[Finetuned via $ACE+ADD^{Inf+foc}$]{\includegraphics[width=0.356\textwidth]{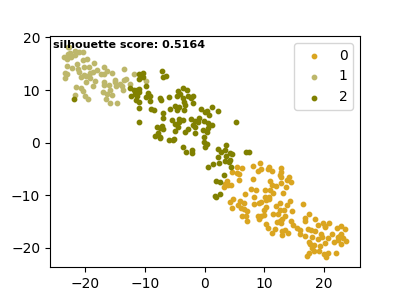}}\vspace{-4mm}
    
    \subfloat[Default BERT]{\includegraphics[width=0.356\textwidth]{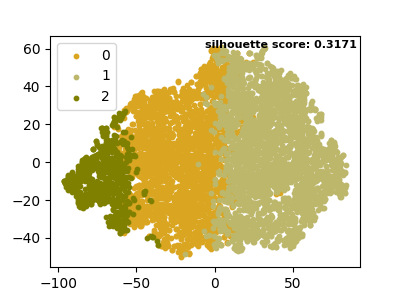}}\hspace{-5.75mm}
    \subfloat[Finetuned via $ACE$]{\includegraphics[width=0.356\textwidth]{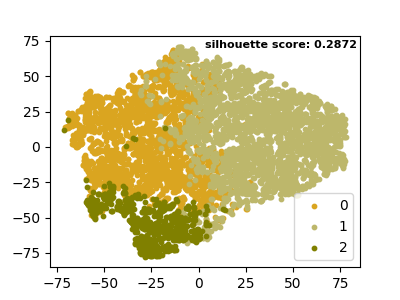}}\hspace{-5.75mm}
    \subfloat[Finetuned via $ACE + ADD^{Inf+foc}$]{\includegraphics[width=0.356\textwidth]{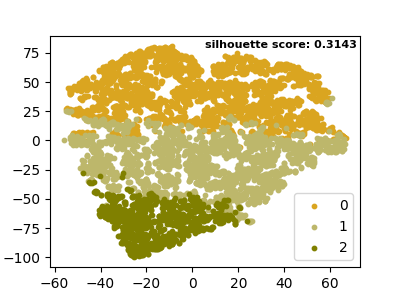}}
    \caption{\textcolor{black}{2D t-SNE plots of the last hidden representations after applying K-means (K=3) on the implicit class for  \abuse\ (a, b, c), \gab\ (d, e, f) and \latent\ (g, h, i). $\{0, 1, 2\}$ are the subcluster ids. The higher the Silhouette score, the better discriminated the clusters.}} 
    \label{fig:vis_add}
\vspace{-3mm}
\end{figure*}

\textbf{Subclustering objective.} After applying the \ADDINFFOC\ objective, we expect not only the per-class clusters to be sufficiently separated but also the subclusters in each class to be better segregated to match their local neighborhood better. Figure \ref{fig:vis_add} shows the implicit embedding space of \abuse, \gab\ and \latent\ after applying K-means on the default BERT embedding (a, d, g), BERT finetuned with the ACE (b, e, h), and \model\ (c, f, i) on three-way hate classification. The higher the Silhouette score, the better the subclusters are separated. $0.34$, $0.31$, and $0.51$  are the scores for cases (a), (b), and (c), respectively, in \abuse. \textcolor{black}{$0.38$, $0.24$, and $0.52$ are the scores for cases (d), (e), and (f), respectively, in \gab. $0.32$, $0.29$, and $0.32$ are the scores for cases (d), (e), and (f), respectively, in \latent.} 

Consequently, an increase of $0.20$, $0.28$, and $0.03$ scores is observed when comparing \model\ with ACE for \abuse, \gab, and \latent\, respectively. This increase in scores validates that the local densities within a class get further refined under \ADDINFFOC\ objective. As expected, ACE sub-optimally treats the implicit class as a single homogeneous cluster. Interestingly, for \latent\, the score does not improve over the default BERT, even though it improves over ACE. A deeper analysis with multiple $K$ values \textcolor{black}{might help here.}

\begin{figure*}[!h]
 \captionsetup[subfigure]{justification=centering}
    \centering
    \subfloat[Default BERT]{\includegraphics[width=0.356\textwidth]{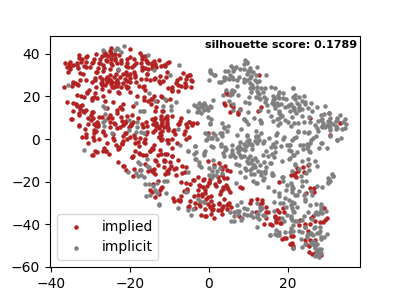}}\hspace{-5.75mm}
    \subfloat[Finetuned via $ACE$]{\includegraphics[width=0.356\textwidth]{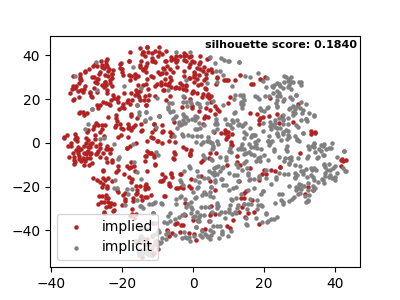}}\hspace{-5.75mm}
    \subfloat[Finetuned with $ACE + ADD^{Inf+foc}$]{\includegraphics[width=0.356\textwidth]{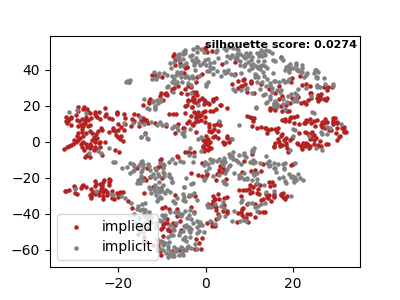}}\vspace{-4mm}  

    \subfloat[Default BERT]{\includegraphics[width=0.356\textwidth]{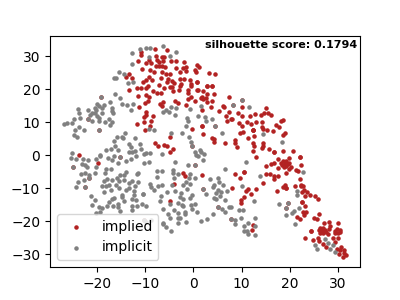}}\hspace{-5.75mm}
    \subfloat[Finetuned via $ACE$]{\includegraphics[width=0.356\textwidth]{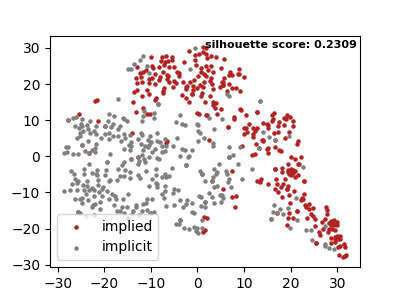}}\hspace{-5.75mm}
    \subfloat[Finetuned with $ACE + ADD^{Inf+foc}$]{\includegraphics[width=0.356\textwidth]{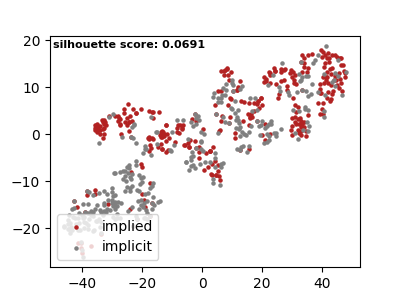}}\vspace{-4mm}
    
    \subfloat[Default BERT]{\includegraphics[width=0.356\textwidth]{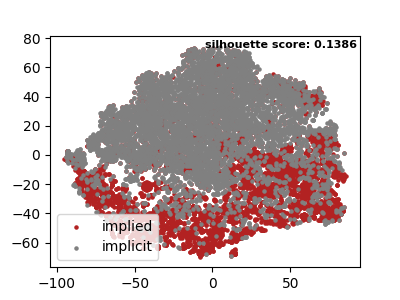}}\hspace{-5.75mm} 
    \subfloat[Finetuned via $ACE$]{\includegraphics[width=0.356\textwidth]{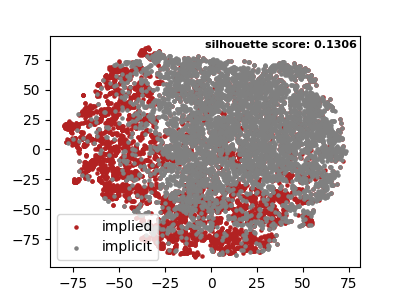}}\hspace{-5.75mm}
    \subfloat[Finetuned with $ACE + ADD^{Inf+foc}$]{\includegraphics[width=0.356\textwidth]{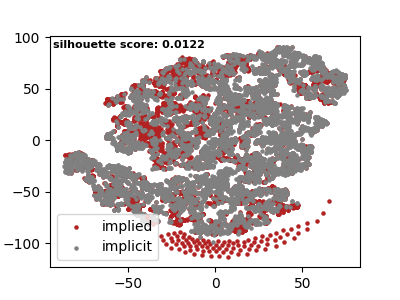}}
    
    \caption{\textcolor{black}{2D t-SNE plots of the last hidden representations obtained for the implicit class and its respective inferential (implied) set for \abuse\ (a, b, c), \gab\ (d, e, f) and \latent\ (g, h, i). The lower the Silhouette score, the closer the surface and implied forms of hate.}}
    \label{fig:vis_inf}
\vspace{-3mm}
\end{figure*}

\textbf{Inferential infusion.} Given that \ADDINFFOC\ brings the surface and semantic forms of implicit hate closer, we expect a significant drop in Silhouette scores between these clusters under \model. Figure \ref{fig:vis_inf}  visualizes the embedding space of default BERT (a, d, g), BERT finetuned with the ACE (b, e, h), and \model\ (c, f, i) on three-way classification for \abuse, \gab, and \latent. \textcolor{black}{$0.18$, $0.18$, and $0.03$ are the scores for cases (a), (b), and (c), respectively, in \abuse. $0.18$, $0.23$, and $0.07$ are the scores for cases (d), (e), and (f), respectively, in \gab. $0.14$, $0.13$, and $0.01$ are the scores for cases (g), (h), and (i), respectively, in \latent.} It is important to highlight that for both BERT and BERT+ACE, there is no explicit objective to bring the implicit and implied clusters together. Hence, they act as a baseline for comparing how well the \ADDINFFOC\ objective brings the two spaces closer. 

A drop of $0.15$, $0.16$, and $0.12$ in the Silhouette score is observed when comparing BERT+ACE with \model\ for \abuse, \gab, and \latent, respectively. It corroborates that the implicit and implied meaning representations are brought significantly closer to each other by employing our model. \textcolor{black}{In addition to Tables \ref{tab:two-way-cls} and \ref{tab:three-way-cls}, the latent space analysis also quantifies our manual annotations for \abuse\ and \gab, as inferential infusion (supported by the manual annotations) is improving the detection of implicit hate.}

\section{Conclusion} 
An increase in hate speech on the Web has necessitated the involvement of automated hate speech detection systems. To this end, we do not recommend completely removing human moderators; instead, we recommend employing machine learning-based systems to perform the first level of filtering. Following the rise of PLMs for text classification, they are now defacto for hate speech detection, too. However, PLM-based systems still suffer from understanding nuanced concepts, such as implicitness, and require external contextualization. 

To this end, \modelfull\ (\model) presents a generalized framework for semantic classification tasks in which the surface form of the source text differs from its inference form. For any system modeling this setup, the aim is to bring the two embedding spaces closer. In this work, the objective is achieved by optimizing for adaptive density discrimination via inferential infusion. Clustering accounts for variation in local neighborhoods beyond a single sample or a single positive/negative pairing; the inferential infusion assures that while we look into the local neighborhood, the implicit clusters are mapped to the apt semantic latent spaces. Further, this work introduces the focal penalty that pays more attention to the sample near the classification boundary. Even by itself, the \ADDFOC\ objective provides a considerable improvement over a standard loss function and can be applied as a substitute.  
 
Overall, our inferential-infused focal \ADDINFFOC\ provides a novel augmentation to the PLM finetuning pipeline. The efficacy of the \model's variants is analyzed over three implicit hate detection datasets (with two of them being manually annotated by us for inferential context), three implicit semantic tasks (sarcasm, irony, and stance detection), and three PLMs (BERT, HateBERT, and XLM). By design, the \ADDINFFOC\ objective does help improve the detection of hate in both two-way and three-way classifications. Our results call into question the role of domain-specific models like HateBERT against BERT as we observe that once finetuned, both of them perform comparably. It calls into question the role of domain-specific models in NLP. 

A more granular examination of \model\ over the latent space for hate speech detection is performed via the analysis of -- seed-wise performance measurement, latent space analysis of the embedding space clusters, and error analysis of positive and negative use cases. Over multiple seeds and 36 experimental setups, we observe the \model\ variants improve over ACE in 32 instances. Meanwhile, a closer look at the later space further highlights the significant improvement that \model\ has on the implicit clusters in bringing them near their implied meaning. 

\section{Limitations and Future Work}
Firstly, the current setup utilizes manual annotations of implicit meaning to be available for inferential clustering, requiring manual effort. Secondly, the proposed setup, being a novel approach in the direction of implicit detection, works on the de facto K-means and uses the same number of subclusters for all datasets. 

In the future, we expect an infusion of generative models \textcolor{black}{to pseudo-annotate the implied meaning, which can be paraphrased and rectified by human annotators on a need basis.} \textcolor{black}{Further, the proposed setup can be employed as an external loss to nudge the LLMs to generate better-quality adversarial examples.} 
Meanwhile, to overcome performing K-means on the entire training set after each epoch, consider representations only for the given batch, starting with stratified sampling so that the batch is representative of the overall dataset. Recent advancements in hashing and dictionary techniques can improve computational efficiency. In the future, we aim to make the system more computationally efficient and extend its application to other tasks. It would be fascinating to review how focal infusion impacts the classification tasks in computer vision in comparison to the \ADD\ setup.

\section*{Ethical Concerns}
This work focuses on textual features and does not incorporate personally identifiable or user-specific signals. For annotations, the annotators were sensitized about the task at hand and given sufficient compensation for their expert involvement. The annotators worked on $\approx$ 250 samples per day over four days to avoid feeling fatigued. Further, the annotators had access to the Web; while annotating, they referred to multiple news sources to understand the context. The dataset of inferential statements for \abuse\ and \gab\ will be available to researchers on request.

\section*{Acknowledgments}
Sarah Masud acknowledges the support of the Prime Minister Doctoral Fellowship in association with Wipro AI and Google India PhD Fellowship. Tanmoy Chakraborty acknowledges the financial support of Anusandhan National Research Foundation (CRG/2023/001351) and Rajiv Khemani Young Faculty Chair Professorship in Artificial Intelligence.

\newpage
\bibliographystyle{nlelike}
\bibliography{nle}
\end{document}